%% file: article.tex
\documentclass[a4paper,twoside]{article}

\usepackage{epsfig}
\usepackage{subcaption}
\usepackage{calc}
\usepackage{amssymb}
\usepackage{amstext}
\usepackage{amsmath}
\usepackage{amsthm}
\usepackage{multicol}
\usepackage{multirow}
\usepackage{tabularx}
\usepackage{xcolor}
\usepackage{colortbl}
\usepackage{pslatex}
\usepackage{apalike}
\usepackage{algorithm2e}
\usepackage{hyperref}
\usepackage[bottom]{footmisc}
\usepackage{SCITEPRESS}     

\newcolumntype{L}[1]{>{\raggedright\let\newline\\\arraybackslash\hspace{0pt}}m{#1}}
\newcolumntype{C}[1]{>{\centering\let\newline\\\arraybackslash\hspace{0pt}}m{#1}}
\newcolumntype{R}[1]{>{\raggedleft\let\newline\\\arraybackslash\hspace{0pt}}m{#1}}

\begin{document}



\title{Large Language Model and Formal Concept Analysis: a comparative study for Topic Modeling}

\author{\authorname{Fabrice BOISSIER\sup{1,2}\orcidAuthor{0000-0002-0067-6524},
Monica SEN\sup{2},
and Irina RYCHKOVA\sup{2}\orcidAuthor{0000-0002-1100-0116}
}
\affiliation{\sup{1}ELSE, EFRITS, 32 Avenue Pierre Semard, Ivry-sur-Seine, France}
\affiliation{\sup{2}CRI, Université Paris 1 Panthéon - Sorbonne, 90 Rue de Tolbiac, Paris, France}
\email{fabrice.boissier@efrits.fr, monica.sen@etu.univ-paris1.fr, irina.rychkova@univ-paris1.fr}
}






\input{article_0_abstract.tex}

\onecolumn \maketitle \normalsize \setcounter{footnote}{0} \vfill


\input{article_1_content}


%



\bibliographystyle{apalike}
{\small
\bibliography{my_bibliography}}


\section*{\uppercase{Appendix}} 
\label{section:X:Appendix}

The Appendix section contains a list of the papers used in the Research Papers dataset.
It also contains metrics and results concerning the CREA and ChatGPT processing of both Abstracts and Full Papers versions of the dataset.

\input{article_2_appendix_papers_list}

\input{article_3_appendix_metrics_clusters}


\end{document}

%% file: article_0_abstract.tex
%


\abstract{
Topic modeling is a research field finding increasing applications: historically from document retrieving, to sentiment analysis and text summarization.
Large Language Models (LLM) are currently a major trend in text processing, but few works study their usefulness for this task.
Formal Concept Analysis (FCA) has recently been presented as a candidate for topic modeling, but no real applied case study has been conducted.
In this work, we compare LLM and FCA to better understand their strengths and weaknesses in the topic modeling field.
FCA is evaluated through the CREA pipeline used in past experiments on topic modeling and visualization, whereas GPT-5 is used for the LLM approach.
A strategy based on three prompts is applied with GPT-5 in a zero-shot setup: topic generation from document batches, merging of batch results into final topics, and topic labeling.
A first experiment reuses the teaching materials previously used to evaluate CREA, while a second experiment analyzes 40 research articles in information systems to compare the extracted topics with the underlying subfields.
}


%% file: article_1_content.tex
\section{\uppercase{Introduction}} 
\label{section:1:Introduction}

Text analysis has been in the spotlight thanks to \textit{Large Language Models} (LLM)~\cite{zhao2025survey}and \textit{ChatGPT}~\cite{openai2022chatgpt}, which has affected practices in numerous domains: from correcting texts~\cite{alsaweed2024investigating} and code \cite{csuvik2023can}, to automatically producing applications from written requirements~\cite{ma2025what}.

\textit{Topic modeling}, a research field dedicated to the extraction and analysis of topics contained within texts, also benefits from this trend~\cite{churchill2022evolution}.
Topic modeling is often performed using \textit{Latent Dirichlet Allocation} (LDA)~\cite{blei2003latent}, a probabilistic approach, \textit{Bidirectional Encoder Representations from Transformers} (BERT)~\cite{devlin2018bert}, a neural-network-based approach or a combination of both~\cite{peinelt2020tbert}\cite{george2023integrated}.
Recent research has proposed \textit{Formal Concept Analysis} (FCA)~\cite{wille2005fca}\cite{stumme2005fca}\cite{ganter2012fca} as a topic modeling candidate~\cite{boissier2024using}.

FCA is a method known for its exactness and transparency, as it finds its roots in the domain of logic.
Data are represented as \textit{objects} constituted of \textit{attributes}.
\textit{Formal concepts} are deduced from the application of a closure operator to find the maximum objects sharing common attributes (and vice versa).
Numerous applications can be found in the literature~\cite{vskopljanac2014formal}\cite{sarmah2015formal}, including topic modeling~\cite{castellanos2017formal}\cite{akhtar2019hierarchical}.
CREA~\cite{boissier2022tel03774087}\cite{boissier2024using} is a pipeline based on FCA and is dedicated to visualizing a corpus of documents, assessing document relevance, and reusing documents to extract their structure and content with the objective of proposing course sessions.

In this paper, we conduct a comparative study between LLM-based and FCA-based methods of topic modeling.
We use GPT-5 and CREA on two datasets to measure their capabilities to extract topics from a corpus of documents.
The first dataset is the one used in previous works on CREA, and the second consists of articles from the authors' prior research in information systems.
Each dataset is thematically linked and coherent, making human validation of the generated topics easier.

This article is organized as follows:
Section~\ref{section:2:Background} recalls what topic modeling is and how to evaluate associated methods, and provides an explanation of FCA, LLM, and their selected implementations.
Section~\ref{section:3:Experiments} presents the two pipelines used in the experimental setup of the comparative study.
Section~\ref{section:4:Results} presents the results of both pipelines on each dataset.
Section~\ref{section:5:Discussion} discusses the advantages and limitations of CREA and GPT-5.
Section~\ref{section:6:Conclusion} concludes the paper.


\section{\uppercase{Background}} 
\label{section:2:Background}

Topic modeling is an unsupervised method of machine learning dedicated to analyzing one or multiple documents and highlighting the main topics addressed~\cite{churchill2022evolution}.
Usually, a topic is a set of \textit{keywords} (or \textit{terms}).
Multiple families of methods aim to extract topics: from probabilistic methods like LDA~\cite{blei2003latent} which calculate the probability of each term being a member of a topic, to embedding-based methods such as Word2Vec~\cite{mikolov2013distributed} or BERT~\cite{devlin2018bert} which consider each term through its surrounding context.
As LLMs are inherently dedicated to language analysis, recent work has examined how well they handle topic modeling ~\cite{wang2023prompting}\cite{li2025large}.
However, as stated in~\cite{li2025large}, no definite answer can be given about which family performs better: LLMs are preferred for their outputs but lack stability and exactness due to hallucinations, whereas traditional topic modeling methods are easier to deploy and people prefer the topics they build themselves~\cite{li2025large}\cite{norton2012ikea}.


\subsection{Formal Concept Analysis through CREA}
\label{subsection:2.1:FCA-CREA}

Formal Concept Analysis~\cite{wille2005fca}\cite{stumme2005fca}\cite{ganter2012fca} is a mathematical method for analyzing and visualizing data through lattices.
The core of FCA manipulates a set of \textit{objects} described by a set of \textit{attributes} to produce a set of \textit{formal concepts}, each combining objects and attributes.
A \textit{formal context} is first built by associating objects and attributes within a binary matrix.
The formal context is then processed through a closure operator to identify all the maximal combinations of objects and attributes, resulting in a list of formal concepts.
The closure is obtained by taking each object, searching all existing combinations with its attributes, then adding another object and repeating the process until all objects are included in a final formal concept.
The generated formal concepts are linked according to their ancestries; for instance, two concepts containing distinct objects become ancestors of the concept containing both, forming a lattice.
The lattice offers a visualization of objects and attributes that highlights relations difficult to observe in the binary matrix.

FCA is called a paradigm as it involves multiple steps in pre- or post-processing.
For instance, the closure operator can be customized to handle nominal or multi-valued input data~\cite{meddouri2020efficient}\cite{souissi2025cnctp}, and different strategies can be applied to binarize the data~\cite{jaffal2015refinement}.
Various metrics can also be calculated from the formal concepts without forming the lattice~\cite{jaffal2016towards}.
Thanks to its capabilities and modularity, FCA has numerous applications~\cite{stumme2005fca}\cite{vskopljanac2014formal}\cite{sarmah2015formal}, including topic modeling~\cite{castellanos2017formal}\cite{akhtar2019hierarchical}.

\begin{figure}
    \centering
    \includegraphics[width=1.0\linewidth]{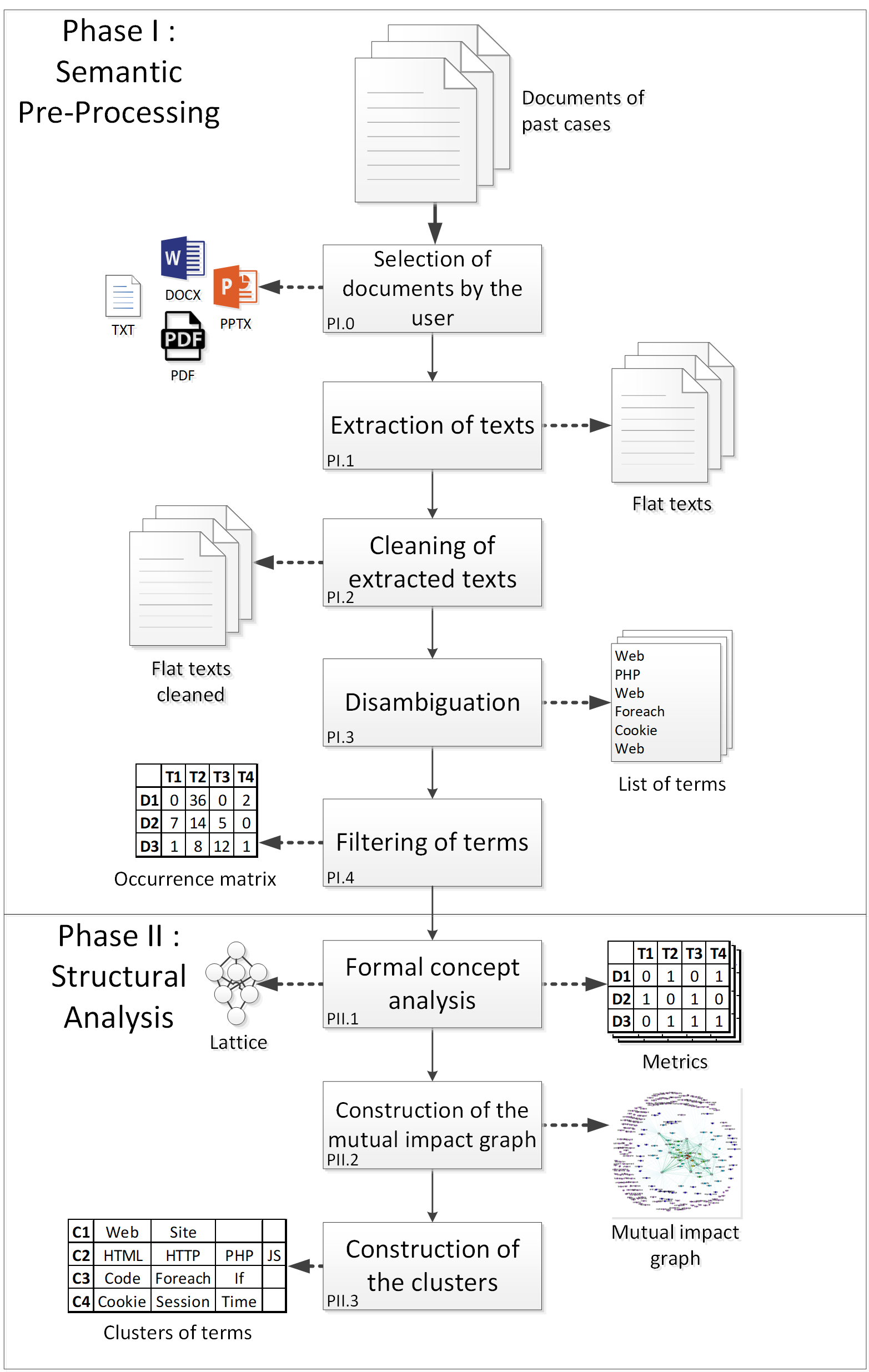}
    \caption{The steps of the CREA pipeline.}
    \label{fig:1:CREA-Pipeline}
\end{figure}

\begin{figure}
    \centering
    \includegraphics[width=1.0\linewidth]{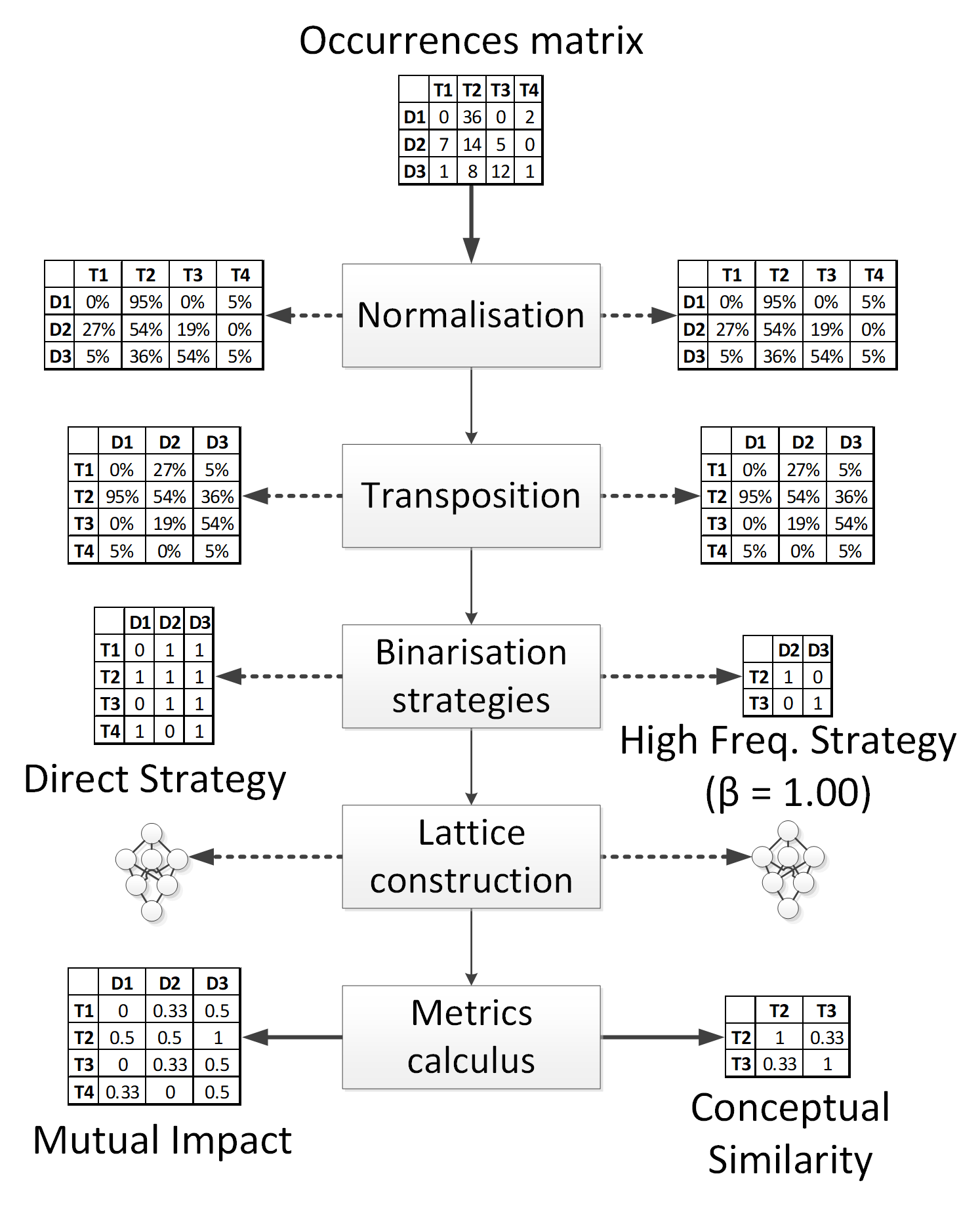}
    \caption{The substeps of the FCA step (PII.1) in the CREA pipeline.}
    \label{fig:2:CREA-FCA-step}
\end{figure}

CREA~\cite{boissier2022tel03774087}\cite{boissier2024using} is a pipeline initially designed for reusing teaching materials by first showing relevance of documents and grouping notions together to organize them as sessions.
Its core relies on BabelFy~\cite{moro2014entity}\cite{moro2014multilingual}, a semantic network designed for identifying named entities, on FCA~\cite{wille2005fca}\cite{stumme2005fca}\cite{ganter2012fca}, for analyzing relations between named entities and their occurrences in documents, and on Hierarchical Agglomerative Clustering (HAC)~\cite{jain1999data} for clustering most relevant notions together within sessions.
Figure~\ref{fig:1:CREA-Pipeline} shows the complete pipeline which generates both the visualization of the relevance through the \textit{Mutual impact graph}, and the topics through the \textit{Clusters of terms}.
The \textbf{Formal concept analysis} (PII.1) step is divided into substeps as illustrated by Figure~\ref{fig:2:CREA-FCA-step}.
The \textbf{binarisation strategies} substep reuses strategies from~\cite{jaffal2015refinement}.
Four strategies are available, and three of them use a threshold ($\beta$) based on term frequencies within the corpora:
\begin{itemize}
    \item \textit{Direct Strategy} turns any number greater than $0$ into a $1$, and any $0$ or negative number into $0$
    \item \textit{Low Strategy} transforms values in the lower frequencies of the term into 1
    \item \textit{High Strategy} transforms values in the higher frequencies of the term into 1
    \item \textit{Medium Strategy} turns values neither in the lower nor the higher frequencies of the term into $1$
\end{itemize}
The \textbf{Metrics calculus} substep also reuses \textit{mutual impact} and \textit{conceptual similarity} from~\cite{jaffal2016towards}.

\subsection{Large Language Models}
\label{subsection:2.2:LLM}

Large Language Models are machine learning models built through self-supervised learning on vast amounts of unlabeled text.
Their specialization in text makes them strong candidates for natural language processing task like topic modeling.
Numerous applications like text clustering~\cite{hoyle2023natural}\cite{zhang2023clusterllm}\cite{viswanathan2024large} or abstractive summarization~\cite{liu2023abstractive} can be found in the literature.
Specifically, as they are pre-trained with a high volume of texts, they better learn the co-occurrence of words in comparison with traditional topic modeling methods~\cite{pham2023topicgpt}.
LLMs complement topic modeling by reducing the need for human interpretation and evaluation of topics~\cite{stammbach2023revisiting}, and can generate topic labels for existing topic models~\cite{rijcken2023towards}.

LLM are used through a prompt providing task instructions to follow.
Multiple strategies of prompting exist like \textit{Zero-shot}~\cite{wang2025mining}
LLMs do have a limitation concerning text length: the context window cannot process large inputs.
In practice, the most common strategy is to split the corpus into subsets, generate topics for each subset, then merge the results into a consolidated list of topics.
In~\cite{doi2024topic} authors called this strategy \textit{parallel prompting}.
They also introduce a variant named \textit{sequential prompting}, which consists of generating topics while taking into account those already identified in previous steps.
Another strategy consists of guiding the LLM by injecting seed topics derived from the original labels into the prompt, with the aim of choosing a desired level of granularity~\cite{pham2023topicgpt}\cite{mu2024large}.
This method constrains the LLM to follow predefined categories.

In~\cite{de2024beyond} the authors state that the more detailed the instructions, the closer LLM-generated topic interpretations align with human annotators.
Their study also shows that LDA achieves a better coherence score (CV) for topic generation, whereas LLMs produce more stable scores.
They conclude that relying on a coherence score alone is insufficient.
This assertion is supported by~\cite{dieng2020topic} and is further extended in~\cite{rijcken2023towards}, which explains that interpretability is a complex notion that cannot be captured by a figure.


\section{\uppercase{Comparative Experiment}} 
\label{section:3:Experiments}

A comparative experiment is a scientific method that aims to compare multiple methods within a controlled framework, in order to measure differences on predefined dependent variables.
The comparison between FCA and LLM is done by producing topics from different datasets and different parameters or prompts.
Three datasets were initially considered, but one was excluded: \textit{20Newsgroups} is often used for assessing topic modeling, but its heterogeneity produced high noise that hindered clear topic identification.
For FCA, we reused the CREA implementation as it already showed topic modeling capability~\cite{boissier2022tel03774087}\cite{boissier2024using}.
For LLM, we first considered Llama 3.1 8B Instruct\footnote{\url{https://huggingface.co/meta-llama/Llama-3.1-8B-Instruct}} because of its open weights; however, it rarely followed prompt instructions and showed high variability in both the format and the structure of its answers.
As a replacement, we followed the literature that reported promising results from OpenAI GPT-5\footnote{\url{https://openai.com/index/introducing-gpt-5/}} in conversational mode.
The experimental protocol follows Pfleeger's recommendations~\cite{pfleeger1995experimental} for designing experiments in software engineering, adapted to topic modeling.


\subsection{Datasets}
\label{subsection:3.1:Datasets}

As stated in~\cite{churchill2022evolution}, assessing topic modeling approaches requires an alignment between their original objectives and the dataset used.
Since CREA was originally applied to teaching materials ~\cite{boissier2022tel03774087}, we reused the same dataset and introduced a second dataset composed of research papers in the information systems field written by the authors.

\paragraph{PHP Courses.}

The dataset is composed of 8 PHP programming courses, which is the corpus used in CREA evaluation~\cite{boissier2022tel03774087}\cite{boissier2024using}.
This dataset is composed of teaching materials in PDF and PowerPoints format that were transformed into raw text with automatic tools, leaving some noise.
The topics contained in the courses concern \textit{web development} using \textit{PHP programming} and \textit{SQL databases}.
The addressed topics are relatively consistent across the whole corpus as they present straightforward way of developing websites.

\paragraph{Research Papers.}

The dataset contains 40 research papers written in english.
The papers concern specific sub-fields of \textit{Information Systems} that are consistent one to another so that extracted topics remain related: \textit{Enterprise Architecture}, \textit{Business Process Management}, and \textit{Information System Engineering}.
Two sub-datasets are generated:
\begin{itemize}
    \item one from the \textbf{abstracts} (without keywords) containing $ 3,343 $ tokens total ($ 2,492 $ unique tokens), with an average of $ 83.6 $ tokens per document and $ 62.3 $ unique tokens per document,
    \item and one from the \textbf{full papers} (without abstracts and keywords) containing $ 223,530 $ tokens total ($ 44,403 $ unique tokens), with an average of $ 5,588 $ tokens per document and $ 1,110 $ unique tokens per document.
\end{itemize}

These datasets concern specific domains and allow evaluating the performance of both FCA and LLM on short and long texts. Both PHP Courses and full papers contain enough text and long sentences to evaluate the relevance and readability of topics generated by both methods on domain-specific data.

These datasets concern specific domains and allow evaluating performance of both FCA and LLM on short and long texts.
Both PHP Courses and Full Papers contain enough text and long sentences to evaluate the relevance and readability of topics generated by both methods on a domain-specific data.


\subsection{Preparation}
\label{subsection:3.2:Preparation}

Before executing any method, we extracted text from documents and applied a minimal cleaning step.
This preparation is required before prompting GPT as non-printable characters and other artifacts create noise that can alter LLM processing and answers.
This step is intentionally minimal in order to avoid excessive cleansing like removing stop words: these words are normally managed by GPT-5, and removing them might also alter its answers.
The PHP Dataset is already extracted\footnote{\href{https://github.com/metalbobinou/CREA-phd-data/tree/main/data/1-extracted-text/PHP-Courses}{https://github.com/metalbobinou/CREA-phd-data}} and only require the minimalist clean step.
The Research Papers Dataset was extracted using PDFPlumber\footnote{\href{https://github.com/jsvine/pdfplumber}{https://github.com/jsvine/pdfplumber}} with custom parameters in order to extract only the text in the different article layouts (one-column, two-columns, ...).
Meta-data are carefully removed (abstract, keywords, and reference section), but tables and figures are kept, even if they become poorly formatted or irrelevant to the surrounding text.


\subsection{CREA Execution}
\label{subsection:3.3:CREA-Execution}

\subsubsection{Semantic Pre-Processing phase}
\label{subsubsection:3.3.1:CREA-Execution-Phase-1}

In the CREA pipeline, the \textbf{Cleaning of extracted texts} (PI.2) step implies a lemmatization and POS (\textit{Part-of-Speech}) filtering processing using TreeTagger~\cite{schmid1994probabilistic}\cite{schmid1995improvements}.
As the original experiment was conducted on French texts, we selected similar POS classes for English in the research papers case.
The Table~\ref{tab:1:CREA-Corpus-Stats} shows the resulting clean statistics on each dataset.
Abstracts have a higher lexical diversity with $ 75 \% $ unique tokens while the two other datasets only show approximately $ 20 \% $ of unique tokens.
Hapax (words appearing exactly only one time in a whole corpus) are even present at $ 60 \% $ within abstracts against $ 10 \% $ and $ 6 \% $ in other corpus.
This can be explained by the fact that abstracts must present succinctly as many keywords as possible, while regular texts manipulate more often the same keywords.

\begin{table*}[!ht]
    \centering
    \caption{Research corpus statistics after lemmatization and POS filtering.}
    \begin{tabular}{ |L{1.35cm}|C{1.35cm}|C{1.35cm}|C{1.35cm}|C{1.35cm}|C{1.35cm}|C{1.35cm}| }
        \hline
        \multirow{3}{*}{\parbox[c]{1.35cm}{Dataset}} &
        \multirow{3}{*}{\parbox[c]{1.25cm}{Total Tokens}} &
        \multirow{3}{*}{\parbox[c]{1.35cm}{Total Unique Tokens}} &
        \multirow{3}{*}{\parbox[c]{1.35cm}{Avg Tokens / Doc}} &
        \multirow{3}{*}{\parbox[c]{1.35cm}{Unique Tokens / Doc}} &
        \multirow{3}{*}{\parbox[c]{1.35cm}{Unique Token Ratio}} &
        \multirow{3}{*}{\parbox[c]{1.35cm}{Hapax ($\%$)}} \\
         & & & & & & \\
         & & & & & & \\
        \hline
        \multirow{2}{*}{Abstracts} & \multirow{2}{*}{$ 3,343 $} & \multirow{2}{*}{$ 2,492 $} & \multirow{2}{*}{$ 84 $} & \multirow{2}{*}{$ 62.3 $} & \multirow{2}{*}{$ 75.2 \% $} & \multirow{2}{*}{$ 60 \% $} \\
         & & & & & & \\
        \hline
        Full & \multirow{2}{*}{$ 223,530 $} & \multirow{2}{*}{$ 44,403 $} & \multirow{2}{*}{$ 5,588 $} & \multirow{2}{*}{$ 1,110 $} & \multirow{2}{*}{$ 20.1 \% $} & \multirow{2}{*}{$ 10 \% $} \\
        Papers & & & & & & \\
        \hline
    \end{tabular}
    \label{tab:1:CREA-Corpus-Stats}
\end{table*}

For the \textbf{Disambiguation} (PI.3) step of CREA, concerning BabelFy~\cite{moro2014entity}\cite{moro2014multilingual}, we reused the parameters as in the previous works: exact matching mode (\textit{EXACT\_MATCHING}) and top-scored candidate selection parameters to keep only the most reliable senses with high (\textit{ScoredCandidates.TOP}) confidence according to BabelFy.

Similarly, in the \textbf{Filtering Terms} (PI.4) step, we kept the terms that have a coherence score above $ 0.05 $, exactly as in previous work.

\subsubsection{Structural Analysis phase}
\label{subsubsection:3.3.2:CREA-Execution-Phase-2}

For the \textbf{Formal concept analysis} (PII.1) step, we produce the formal concepts using the python library \textit{concepts}~\cite{bank2016concepts}~\footnote{\url{https://github.com/xflr6/concepts}}.
In our tests, since we used far more texts than the original experiments in~\cite{boissier2022tel03774087}\cite{boissier2024using}, we followed their conclusions by using values of $ \beta > 1.00 $ in the \textbf{Binarisation strategies} substep.
Next, in the \textbf{Metrics calculus} substep, we applied only the \textit{conceptual similarity} in order to produce the topics (the relevance visualization being out of scope).

Table~\ref{tab:2:CREA-Strategy-UniqueTerms-Abstracts} shows the number of unique terms kept for each strategy on the abstracts, and Table~\ref{tab:3:CREA-Strategy-Concepts-Abstracts} shows the number of formal concepts generated.
High and low strategies retain less than $40\%$ of terms and even less than $ 50 $ terms at $\beta \geq 1.00$, while medium strategy retains full vocabulary and a manageable count of formal concepts at $ 0.75 \leq \beta \leq 1.00 $.

\begin{table}[!ht]
    \centering
    \caption{Number of unique terms per strategy on Abstracts.}
    \begin{tabular}{ | c |c|c|c|c| }
        \hline
         \multirow{2}{*}{$ \beta $} &
        \multicolumn{4}{c|}{Strategy} \\
        \cline{2-5}
         & Direct & Medium & High & Low \\
        \hline
        $ 0.00 $ & \multirow{9}{*}{ $ 470 $ } & $ 295 $ & $ 175 $ & $ 175 $ \\
        \cline{1-1} \cline{3-5}
        $ 0.25 $ & & $ 347 $ & $ 175 $ & $ 175 $ \\
        \cline{1-1} \cline{3-5}
        $ 0.50 $ & & $ 377 $ & $ 175 $ & $ 175 $ \\
        \cline{1-1} \cline{3-5}
        $ 0.75 $ & & $ 386 $ & $ 175 $ & $ 169 $ \\
        \cline{1-1} \cline{3-5}
        $ 1.00 $ & & $ 470 $ & $  91 $ & $  84 $ \\
        \cline{1-1} \cline{3-5}
        $ 1.25 $ & & $ 470 $ & $  76 $ & $  29 $ \\
        \cline{1-1} \cline{3-5}
        $ 1.50 $ & & $ 470 $ & $  51 $ & $   3 $ \\
        \cline{1-1} \cline{3-5}
        $ 1.75 $ & & $ 470 $ & $  27 $ & $   0 $ \\
        \cline{1-1} \cline{3-5}
        $ 2.00 $ & & $ 470 $ & $  17 $ & $   0 $ \\
        \hline
    \end{tabular}
    \label{tab:2:CREA-Strategy-UniqueTerms-Abstracts}
\end{table}

\begin{table}[!ht]
    \centering
    \caption{Number of generated concepts per strategy on Abstracts.}
    \begin{tabular}{ | c |c|c|c|c| }
        \hline
         \multirow{2}{*}{$ \beta $} &
        \multicolumn{4}{c|}{Strategy} \\
        \cline{2-5}
         & Direct & Medium & High & Low \\
        \hline
        $ 0.00 $ & \multirow{9}{*}{ $ 558 $ } & $  43 $ & $ 119 $ & $ 191 $ \\
        \cline{1-1} \cline{3-5}
        $ 0.25 $ & & $  68 $ & $  87 $ & $ 144 $ \\
        \cline{1-1} \cline{3-5}
        $ 0.50 $ & & $  98 $ & $  70 $ & $ 107 $ \\
        \cline{1-1} \cline{3-5}
        $ 0.75 $ & & $ 153 $ & $  61 $ & $  63 $ \\
        \cline{1-1} \cline{3-5}
        $ 1.00 $ & & $ 255 $ & $  49 $ & $  37 $ \\
        \cline{1-1} \cline{3-5}
        $ 1.25 $ & & $ 361 $ & $  39 $ & $  20 $ \\
        \cline{1-1} \cline{3-5}
        $ 1.50 $ & & $ 448 $ & $  29 $ & $   5 $ \\
        \cline{1-1} \cline{3-5}
        $ 1.75 $ & & $ 500 $ & $  20 $ & $   5 $ \\
        \cline{1-1} \cline{3-5}
        $ 2.00 $ & & $ 534 $ & $  14 $ & $   5 $ \\
        \hline
    \end{tabular}
    \label{tab:3:CREA-Strategy-Concepts-Abstracts}
\end{table}

For the full papers, the high strategy is used because the vocabulary is much larger and therefore requires filtering.
The high strategy has already been used with this objective in previous works~\cite{boissier2022tel03774087}\cite{boissier2024using}.
The Table~\ref{tab:4:CREA-Strategy-UniqueTerms-Concepts-FullPapers} shows the unique terms and concepts for the direct and high strategies.
A large amount of vocabulary is preserved with $ 0.75 \leq \beta \leq 1.25 $, but many formal concepts are also produced.
In contrast, $ \beta \geq 3.00 $ excludes nearly $ 97\% $ of the vocabulary, making it too narrow for a realistic analysis.

\begin{table}[!ht]
    \centering
    \caption{Number of unique terms and concepts per strategy on Full Papers.}
    \begin{tabular}{ | c |c|c|c|c| }
        \hline
        \multirow{4}{*}{$ \beta $} &
        \multicolumn{4}{c|}{Strategy} \\
        \cline{2-5}
         & Direct & \multicolumn{3}{c|}{High} \\
        \cline{2-5}
         & Terms & \multirow{2}{*}{Terms} & Excluded & \multirow{2}{*}{Concepts} \\
         & (Concepts) & & ($\%$) & \\
        \hline
        $ 0.75 $ & \multirow{14}{*}{ \shortstack{ $ 3110 $ \\ ($ 9535 $) } } & $ 1602 $ & $ 48.49 $ & $ 666 $ \\
        \cline{1-1} \cline{3-5}
        $ 1.00 $ & & $ 1040 $ & $ 66.56 $ & $ 406 $ \\
        \cline{1-1} \cline{3-5}
        $ 1.25 $ & & $  906 $ & $ 70.87 $ & $ 263 $ \\
        \cline{1-1} \cline{3-5}
        $ 1.50 $ & & $  625 $ & $ 79.90 $ & $ 175 $ \\
        \cline{1-1} \cline{3-5}
        $ 1.75 $ & & $  453 $ & $ 85.43 $ & $ 126 $ \\
        \cline{1-1} \cline{3-5}
        $ 2.00 $ & & $  343 $ & $ 88.97 $ & $  96 $ \\
        \cline{1-1} \cline{3-5}
        $ 2.25 $ & & $  248 $ & $ 92.03 $ & $  71 $ \\
        \cline{1-1} \cline{3-5}
        $ 2.50 $ & & $  176 $ & $ 94.34 $ & $  48 $ \\
        \cline{1-1} \cline{3-5}
        $ 2.75 $ & & $  122 $ & $ 96.08 $ & $  44 $ \\
        \cline{1-1} \cline{3-5}
        $ 3.00 $ & & $   78 $ & $ 97.49 $ & $  39 $ \\
        \cline{1-1} \cline{3-5}
        $ 3.25 $ & & $   52 $ & $ 98.33 $ & $  32 $ \\
        \cline{1-1} \cline{3-5}
        $ 3.50 $ & & $   37 $ & $ 98.81 $ & $  24 $ \\
        \cline{1-1} \cline{3-5}
        $ 3.75 $ & & $   24 $ & $ 99.23 $ & $  19 $ \\
        \cline{1-1} \cline{3-5}
        $ 4.00 $ & & $   15 $ & $ 99.52 $ & $  12 $ \\
        \hline
    \end{tabular}
    \label{tab:4:CREA-Strategy-UniqueTerms-Concepts-FullPapers}
\end{table}

Based on vocabulary variety and the number of formal concepts generated, we consider these as the best configurations:
\begin{itemize}
    \item Abstracts: \textit{medium strategy} with $ 0.75 \leq \beta \leq 1.00 $ (complemented by \textit{high strategy} for comparison).

    \item Full Papers: \textit{high strategy} with $ \beta = 1.50 $ as the main configuration (625 terms, 175 concepts), and $ \beta = 2.50 $ for aggressive filtering comparison (176 terms, 48 concepts).
\end{itemize}

\begin{table*}[!ht]
    \centering
    \caption{Cluster metrics using \textit{Medium Strategy} ($\beta = 1.25$) on PHP dataset.}
    \begin{tabular}{ | >{\columncolor{black!15}}c | c|c|c|c| c|c|c|c| }
    \hline
    \rowcolor{black!15} \textit{k} &
    Silh.($\uparrow$) &
    CHI($\uparrow$) &
    DI($\uparrow$) &
    DBI($\downarrow$) &
    min &
    max &
    BalanceRatio &
    Largest ($\%$) \\
    \hline
    $2$ & $0.25$ & $ 8.79$ & $0.63$ & $1.72$ & $13$ & $16$ & $0.81$ & $55.17$ \\ \hline
    $3$ & $0.33$ & $ 9.14$ & $0.63$ & $1.22$ & $6$ & $16$ & $0.38$ & $55.17$ \\ \hline
    $4$ & $0.39$ & $10.46$ & $0.65$ & $1.24$ & $6$ & $9$ & $0.67$ & $31.03$ \\ \hline
    $5$ & $0.46$ & $12.76$ & $0.79$ & $0.91$ & $3$ & $7$ & $0.43$ & $24.14$ \\ \hline
    $6$ & $0.56$ & $14.16$ & $0.79$ & $0.76$ & $3$ & $7$ & $0.43$ & $24.14$ \\ \hline
    $7$ & $0.62$ & $16.49$ & $0.83$ & $0.70$ & $3$ & $6$ & $0.50$ & $20.69$ \\ \hline
    $8$ & $0.68$ & $18.96$ & $0.86$ & $0.61$ & $2$ & $6$ & $0.33$ & $20.69$ \\ \hline
    $9$ & $0.73$ & $22.32$ & $0.86$ & $0.70$ & $2$ & $4$ & $0.50$ & $13.79$ \\ \hline
    $10$ & $0.74$ & $25.47$ & $0.87$ & $0.65$ & $1$ & $4$ & $0.25$ & $13.79$ \\ \hline
    \end{tabular}
    \label{tab:5:CREA-Cluster-Metrics}
\end{table*}

\subsubsection{Clustering phase}
\label{subsubsection:3.3.3:CREA-Execution-Phase-3}

The original experiments in~\cite{boissier2022tel03774087}\cite{boissier2024using} aimed to build a course divided in 8 sessions, requiring to create 8 clusters.
In this paper, we consider $ k $ (the number of clusters) to be a customizable parameter exactly like $ \beta $.
We tested values of $ k $ from $ 2 $ to the number of documents ($ 40 $) by steps of 2.
To choose the best $ k $, we used two sets of metrics dedicated to clustering quality, specifically to improve cluster separation while avoiding dominant clusters or singletons.
These metrics are implemented using \textit{scikit-learn}~\cite{scikitlearn}.
Initially, our evaluation framework was designed to produce multiple standard topic modeling metrics (e.g., coherence measures such as CV, NPMI, and UMass). However preliminary experiments revealed strong imbalances in CREA's outputs, particularly in topic size.
These asymmetries made conventional topic coherence measures less informative, leading us follow the choices made in~\cite{castellanos2017formal} who applied clustering validity indices in the context of FCA for topic detection.


\noindent \textbf{Cluster-structure metrics}:
\begin{itemize}
    \item \textit{Minimum cluster size (min)}

    \item \textit{Maximum cluster size (max)}

    \item \textit{Balance ratio} ($ \text{\textit{BalanceRatio}} = \dfrac{\text{size}_{min}}{\text{size}_{max}} $)

    \item \textit{Percentage of the largest cluster (Largest $\%$)}
\end{itemize}


\noindent \textbf{Internal validity metrics}:
\begin{itemize}
    \item \textit{Silhouette Coefficient (Silh.)}: compares how close each point is to its own cluster versus other clusters (between -1 and 1; closer to 1 is better).

    \item \textit{Calinski-Harabasz Index (CHI)}: measures the ratio of between-cluster variance to within-cluster variance (higher is better).

    \item \textit{Dunn Index (DI)}: evaluates the ratio of intra-cluster to inter-cluster distances (higher is better).

    \item \textit{Davies-Bouldin Index (DBI)}: also evaluates the ratio of intra-cluster to inter-cluster distances (lower is better).
\end{itemize}

Table~\ref{tab:5:CREA-Cluster-Metrics} shows the metrics obtained for different $ k $ values.
Moving from $ k = 2 $ to $ k = 5 $ shows improved internal validity scores (Silhouette: $0.46$, Calinski-Harabasz: $12.76$, Dunn: $0.79$, Davies-Bouldin: $0.91$).
We note that the dominant cluster disappears as its proportion decreases from $ 55\% $ to $ 24\% $.
Cluster balance becomes acceptable with sizes ranging from $ 3 $ to $ 7 $ terms, with a ratio of $ 0.43 $.
As $ k $ increases, clustering metrics continue to improve but at the cost of increased fragmentation.
At $ k = 10 $, scores reach their peak but introduce a singleton cluster ($ min = 1 $), indicating over-segmentation.
Based on the resulting clusters shown in Table~\ref{tab:5:CREA-Cluster-Metrics}, $ k = 5 $ provides a reasonable trade-off between metric optimization and interpretable cluster structure for this example.

Finally, for the \textbf{Hierarchical agglomerative clustering} (HAC), we kept the parameters from the previous work, excepted that the number of clusters was determined by the previous metrics.

\begin{flushleft}
\begin{enumerate}
    \item Pre-processing: \linebreak
    \textit{sklearn.preprocessing.scale()}

    \item Aggregation of clusters: \linebreak
    \textit{scipy.cluster.hierarchy.linkage()} \linebreak
    Parameters: \linebreak
    \textit{method = 'ward', metric = 'euclidean'}

    \item Extraction of clusters with: \linebreak
    \textit{scipy.cluster.hierarchy.fcluster()} \linebreak
    Parameters: \linebreak
    \textit{criterion = 'maxclust', nb\_clusters = k}
\end{enumerate}
\end{flushleft}


\subsection{GPT-5 Execution}
\label{subsection:3.4:GPT5-Execution}

\subsubsection*{Prompting Strategies}
\label{subsubsection:3.4.1:GPT5-Prompt-Strategies}

To handle context length limitations, we adopted a batch-processing strategy inspired by existing approaches~\cite{pham2023topicgpt}\cite{doi2024topic}\cite{de2025llm}.
The process involved three steps:
\begin{flushleft}
\begin{enumerate}
    \item Topic generation from document batches \linebreak
    (see Table~\ref{tab:6:GPT5-Prompt-TopicGeneration} for the prompt template)

    \item Merging of batch results into final topics \linebreak
    (see Table~\ref{tab:7:GPT5-Prompt-TopicsMerging} for the prompt template)

    \item Topic labeling \linebreak
    (see Table~\ref{tab:8:GPT5-Prompt-TopicLabeling} for the prompt template)
\end{enumerate}
\end{flushleft}

To remain within the model's context window, batch sizes ranged from 1 to 10 documents depending on corpus length.
For each batch, we requested five topics expressed in five words.
This procedure corresponds to a zero-shot setup, testing the model's ability to perform topic modeling without any prior examples.

\begin{center}
\begin{table}[!ht]
    \centering
    \caption{Template of prompt for topic generation.}
    \begin{tabular}{ | L{0.43\textwidth} | }
    \hline
    You are simulating a topic modeling system. \\
    Analyze the following set of documents and identify 5 topics. \\
    \textit{ \lbrack DOCUMENTS\rbrack } \\
    Each topic must be represented only by 5 keywords (1–2 words each). \\
    Return solely the results in the following format and nothing else: \\
    Topic k: word, word, word, word, word \\
    ... \\
    \hline
    \end{tabular}
    \label{tab:6:GPT5-Prompt-TopicGeneration}
\end{table}
\end{center}

\begin{center}
\begin{table}[!ht]
    \centering
    \caption{Template of prompt for merging outputs of batch into final topics.}
    \begin{tabular}{ | L{0.43\textwidth} | }
    \hline
    You are consolidating topic modeling results from multiple document batches. \\
    Each batch produced topics in the format "Topic k: word, word, word, word, word". \\
    Here are the topics: \\
    \textit{ \lbrack BATCH RESULTS\rbrack } \\
    Merge these results into exactly 5 final topics. \\
    - Each topic must contain exactly 5 keywords. \\
    - Keywords must be 1–2 words each. \\
    - Merge duplicates and synonyms into a single topic. \\
    - Favor topics that appear in multiple batches. \\
    - Discard topics that appear rarely. \\
    Return solely the final topics in the following format and nothing else: \\
    Topic k: word, word, word, word, word \\
    ... \\
    \hline
    \end{tabular}
    \label{tab:7:GPT5-Prompt-TopicsMerging}
\end{table}
\end{center}

\begin{center}
\begin{table}[!ht]
    \centering
    \caption{Template of prompt for labeling topics.}
    \begin{tabular}{ | L{0.43\textwidth} | }
    \hline
    You are labeling the final topics obtained from topic modeling. \\
    Here are the topics: \\
    \textit{ \lbrack TOPICS\rbrack } \\
    For each topic: \\
    - Assign a concise label (1–2 words). \\
    - Provide a one-sentence description that summarizes the topic. \\
    - Do not add explanations or commentary. \\
    Return solely the final labels in the following format and nothing else: \\
    Topic k : Label - Description \\
    ... \\
    \hline
    \end{tabular}
    \label{tab:8:GPT5-Prompt-TopicLabeling}
\end{table}
\end{center}


\section{\uppercase{Results}} 
\label{section:4:Results}

We present in this section only the most important results for readability issue, but complete results can be found in the~\hyperref[section:X:Appendix]{appendix section}.

\begin{center}
\begin{table*}[!ht]
    \centering
    \caption{Topics and their ChatGPT labels for CREA on PHP dataset using \textit{High Strategy} ($\beta = 1.00$) and $k = 8$.}
    \begin{tabular}{ | c | L{14.0cm} | }
    \hline
    \cellcolor{black!15} Topic & Label / \cellcolor{black!15} Terms \\
    \hline
    \multirow{2}{*}{1}
    & \cellcolor{red!5} PHP Syntax - Basic PHP coding constructs including loops, arrays, classes, and database functions \\ \cline{2-2}
    & \textit{php, code, fois, post, jour, foreach, cle, classe, class, mysqli} \\ \hline
    \multirow{2}{*}{2}
    & \cellcolor{red!5} Web Pages - Interaction between web pages, browsers, servers, and users with associated content \\ \cline{2-2}
    & \textit{page web, navigateur, serveur web, texte, concerner, délimiter, utilisateur, associer, personne, machine, mysql} \\ \hline
    \multirow{2}{*}{3}
    & \cellcolor{red!5} Sessions \& URLs - Handling of URLs, sessions, headers, and access control in web applications \\ \cline{2-2}
    & \textit{url, langage, case, fermeture, session, chaîne, entête, avoir accès} \\ \hline
    \multirow{2}{*}{4}
    & \cellcolor{red!5} Files \& Forms - Managing files, comments, checkboxes, and client-server interpretation \\ \cline{2-2}
    & \textit{fichier, commentaire, case à cocher, interpréter, côté serveur, serveur, côté client} \\ \hline
    \multirow{2}{*}{5}
    & \cellcolor{red!5} Data Types - Typing, keywords, transactions, and displaying information for visitors \\ \cline{2-2}
    & \textit{typage, mot, moteur, affiche, transaction, visiteur} \\ \hline
    \multirow{2}{*}{6}
    & \cellcolor{red!5} Databases - Database operations with insert statements, varchar fields, and null values \\ \cline{2-2}
    & \textit{base de données, insert, varchar, null} \\ \hline
    \multirow{2}{*}{7}
    & \cellcolor{red!5} XML \& Config - Use of XML, configuration files, composer, and document type declarations \\ \cline{2-2}
    & \textit{xml, configuration, composer, doctype} \\ \hline
    \multirow{2}{*}{8}
    & \cellcolor{red!5} Web Data - Handling data, POST methods, scripting languages, timestamps, and file inputs \\ \cline{2-2}
    & \textit{donnée, text, méthode post, programmation, site, langage de script, list, méthode, timestamp, files} \\ \hline
    \end{tabular}
    \label{tab:9:Results-CREA-labels-GPT-DatasetPHP}
\end{table*}
%
%
\begin{table*}[!ht]
    \centering
    \caption{Topics and labels generated with ChatGPT from PHP dataset ($k = 8$).}
    \begin{tabular}{ | c | L{14.0cm} | }
    \hline
    \cellcolor{black!15} Topic & \cellcolor{black!15} Label / Terms \\
    \hline
    \multirow{2}{*}{1}
    & \cellcolor{red!5} Basics - PHP installation, configuration, and basic syntax \\ \cline{2-2}
    & \textit{php, installation, configuration, syntax, echo} \\ \hline
    \multirow{2}{*}{2}
    & \cellcolor{red!5} Variables - Data types, arrays, operators, and constants in PHP \\ \cline{2-2}
    & \textit{variables, types, arrays, operators, constants} \\ \hline
    \multirow{2}{*}{3}
    & \cellcolor{red!5} Control Flow - Conditional statements, loops, and function structures \\ \cline{2-2}
    & \textit{conditionals, loops, control, switch, functions} \\ \hline
    \multirow{2}{*}{4}
    & \cellcolor{red!5} Forms - User input handling with GET/POST and validation \\ \cline{2-2}
    & \textit{forms, input, GET, POST, validation} \\ \hline
    \multirow{2}{*}{5}
    & \cellcolor{red!5} Sessions - Session handling, cookies, login, and security persistence \\ \cline{2-2}
    & \textit{sessions, cookies, login, persistence, csrf} \\ \hline
    \multirow{2}{*}{6}
    & \cellcolor{red!5} OOP - Object-oriented programming with classes, objects, and methods \\ \cline{2-2}
    & \textit{classes, objects, inheritance, methods, attributes} \\ \hline
    \multirow{2}{*}{7}
    & \cellcolor{red!5} Database - SQL queries, PDO, MySQL connections, and transactions \\ \cline{2-2}
    & \textit{database, mysql, queries, pdo, transactions} \\ \hline
    \multirow{2}{*}{8}
    & \cellcolor{red!5} Frameworks - MVC architecture with Symfony, Twig, REST, and related tools \\ \cline{2-2}
    & \textit{mvc, symfony, twig, rest, framework} \\ \hline
    \end{tabular}
    \label{tab:10:Results-GPT-DatasetPHP}
\end{table*}
\end{center}


\subsection{PHP Courses Results}
\label{subsection:4.1:Dataset-PHPCourses-Results}

Concerning the PHP Dataset, we reused the original data from CREA with 8 clusters (for the 8 sessions), but, we also asked ChatGPT to produce a label for each topic.
The model was able to adapt to French terms, and produced correct labeling as shown in Table~\ref{tab:9:Results-CREA-labels-GPT-DatasetPHP}.
When asked for generating topics from the documents, ChatGPT followed correctly the instructions and produced more uniform topics.
It even produced a more logical progression between topics (from basics to advanced notions) as shown in Table~\ref{tab:10:Results-GPT-DatasetPHP}.

Several limitations appear in CREA-generated topics.
Topic size is imbalanced: Topic 2 contains 11 terms, while Topic 6 has only 4.
Redundancies are frequent: Topic 1 includes both \textit{fois} and \textit{foreach}, as well as \textit{classe} and \textit{class}.
Topic 4 contains both \textit{côté serveur} and \textit{serveur}, creating conceptual overlap.
Thematic focus is sometimes unclear: Topic 2 mixes different elements of the web ecosystem (\textit{navigateur}, \textit{serveur web}, \textit{utilisateur}, \textit{personne}, \textit{machine}).
Although these terms may be familiar to someone familiar with web development, it lacks a coherent thematic scope.
Granularity is inconsistent: database-related terms are scattered across topics, with \textit{mysqli} in Topic 1, while \textit{base de données}, \textit{insert}, \textit{varchar}, and \textit{null} are grouped in Topic 6.
These terms range from general database concepts to specific SQL keywords, which complicates interpretation.
However, it must be reminded that the extracted topics are built with the objective of proposing sessions of courses from teaching materials, which implies a certain overlapping of notions.

When we compare the concepts from both methods, we see that ChatGPT introduces structured pedagogical notions such as \textit{installation}, \textit{configuration}, \textit{syntax}, \textit{operators}, \textit{constants}, \textit{validation}, and modern frameworks like \textit{MVC}, \textit{Symfony}, and \textit{REST}, which do not appear in CREA topics.
These terms reflect a learning path from initial setup to advanced concepts.
CREA instead captures more technical details such as \textit{mysqli}, \textit{varchar}, \textit{null}, \textit{composer}, and \textit{doctype}, as well as architectural distinctions like \textit{côté client}, \textit{côté serveur}, and \textit{entête}.
ChatGPT follows familiar educational patterns, whereas CREA's raw terms might help create new courses.


\subsection{Research Papers Results}
\label{subsection:4.2:Dataset-ResearchPapers-Results}

\subsubsection{Abstracts}
\label{subsubsection:4.2.1:Dataset-ResearchPapers-Results-Abstracts}

For abstracts, we selected the medium strategies with $ \beta = 0.75 $, $ \beta = 1 $,  $ \beta = 1.25 $, as these parameters preserved the vocabulary.
We also included the high strategy to compare the effect of reducing the number of concepts.
Table~\ref{tab:11:Results-CREA-Abstracts-TermsConcepts} recalls the number of terms and concepts on Abstracts with high and medium strategy.

\begin{table}[!ht]
    \centering
    \caption{Number of terms and concepts on Abstract with high strategy.}
    \begin{tabular}{ | c |c|c|c|c| }
        \hline
        \multirow{2}{*}{$ \beta $} &
        \multicolumn{2}{c|}{High Strategy} &
        \multicolumn{2}{c|}{Medium Strategy} \\
        \cline{2-5}
         & Terms & Concepts & Terms & Concepts \\
        \hline
        $ 0.75 $ & $ 175 $ & $ 61 $ & $ 386 $ & $ 153 $ \\ \hline
        $ 1.00 $ & $  91 $ & $ 49 $ & $ 470 $ & $ 255 $ \\ \hline
        $ 1.25 $ & $  76 $ & $ 39 $ & $ 470 $ & $ 361 $ \\ \hline
    \end{tabular}
    \label{tab:11:Results-CREA-Abstracts-TermsConcepts}
\end{table}

For simplicity, we chose three different $ k $ values to illustrate the results: $ k = 8 $, $ k = 16 $, and $ k = 20 $ (see Table~\ref{tab:12bis:Results-CREA-Abstracts-Clusters-Size-K=8-16-20}).
The cluster size distributions consistently reveal an imbalance across all configurations, with one dominant cluster systematically absorbing the majority of terms while the second largest cluster remains disproportionately small.
For example, with $ \beta = 1 $, which corresponds to complete vocabulary conservation with 255 concepts, the largest cluster contains 361 terms.
When applying the high strategy with the same $ \beta $ value, the filtering reduces the vocabulary to a more manageable 49 concepts for 91 terms, yet the imbalance persists with the largest cluster containing 55 terms while the second largest contains only 8 terms.
We observe this pattern across all $ \beta $ values, suggesting that one cluster tends to absorb the majority of terms regardless of the filtering strategy employed.

\begin{table*}[!ht]
    \centering
    \caption{Cluster size on Abstract across $ \beta $, with \textit{High} and \textit{Medium Strategy}, and $ k = 8, 16, 20 $.}
    \begin{tabular}{ | c | c | c|c|c| c|c|c| }
        \hline
        \multirow{2}{*}{$ k $} &
        \multirow{2}{*}{$ \beta $} &
        \multicolumn{3}{c|}{High Strategy} &
        \multicolumn{3}{c|}{Medium Strategy} \\
        \cline{3-8}
         & &
        Largest & $2^{\text{nd}}$ cluster & Smallest &
        Largest & $2^{\text{nd}}$ cluster & Smallest \\
        \hline
        \multirow{3}{*}{$ k = 8 $}
         & $ 0.75 $ & $ 122 $ & $ 11 $ & $ 6 $ & $ 290 $ & $ 16 $ & $ 12 $\\ \cline{2-8}
         & $ 1.00 $ & $  55 $ & $  8 $ & $ 4 $ & $ 361 $ & $ 27 $ & $ 12 $ \\ \cline{2-8}
         & $ 1.25 $ & $  41 $ & $  7 $ & $ 3 $ & $ 370 $ & $ 18 $ & $ 12 $ \\
        \hline \hline
        \multirow{3}{*}{$ k = 16 $}
         & $ 0.75 $ & $  73 $ & $ 11 $ & $ 4 $ & $ 202 $ & $ 16 $ & $ 10 $\\ \cline{2-8}
         & $ 1.00 $ & $  24 $ & $  8 $ & $ 3 $ & $ 270 $ & $ 21 $ & $ 11 $ \\ \cline{2-8}
         & $ 1.25 $ & $  17 $ & $  7 $ & $ 2 $ & $ 247 $ & $ 34 $ & $ 10 $ \\
        \hline \hline
        \multirow{3}{*}{$ k = 20 $}
         & $ 0.75 $ & $  57 $ & $ 11 $ & $ 4 $ & $ 163 $ & $ 16 $ & $ 9 $\\ \cline{2-8}
         & $ 1.00 $ & $  16 $ & $  8 $ & $ 2 $ & $ 196 $ & $ 44 $ & $ 10 $ \\ \cline{2-8}
         & $ 1.25 $ & $   9 $ & $  7 $ & $ 2 $ & $ 210 $ & $ 34 $ & $ 9 $ \\
        \hline
    \end{tabular}
    \label{tab:12bis:Results-CREA-Abstracts-Clusters-Size-K=8-16-20}
\end{table*}

\begin{figure}[!ht]
    \centering
    \includegraphics[width=1.0\linewidth]{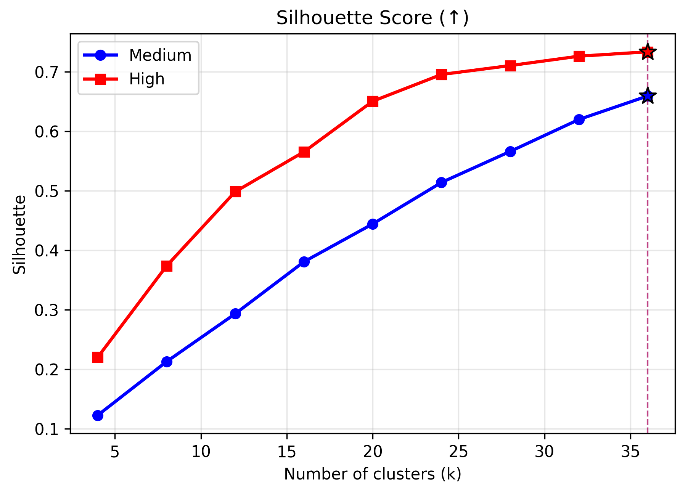}
    \caption{Silhouette score on clusters from Abstracts concerning \textit{Medium} vs. \textit{High Strategy} ($ \beta = 1.00 $)}
    \label{fig:3.1:Results-CREA-Abstracts-ClustersMetrics-Silhouette}
\end{figure}

\begin{figure}[!ht]
    \centering
    \includegraphics[width=1.0\linewidth]{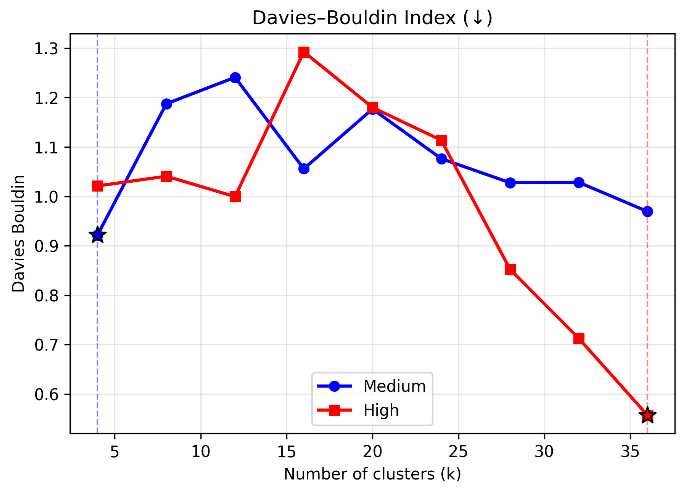}
    \caption{Davies-Bouldin index on clusters from Abstracts concerning \textit{Medium} vs. \textit{High Strategy} ($ \beta = 1.00 $)}
    \label{fig:3.2:Results-CREA-Abstracts-ClustersMetrics-DB}
\end{figure}

\begin{figure}[!ht]
    \centering
    \includegraphics[width=1.0\linewidth]{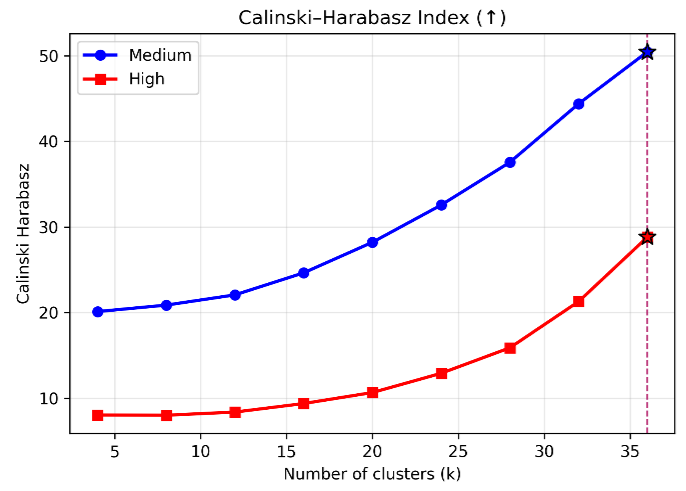}
    \caption{Calinski-Harabasz index on clusters from Abstracts concerning \textit{Medium} vs. \textit{High Strategy} ($ \beta = 1.00 $)}
    \label{fig:3.3:Results-CREA-Abstracts-ClustersMetrics-CH}
\end{figure}

Figures~\ref{fig:3.1:Results-CREA-Abstracts-ClustersMetrics-Silhouette}-\ref{fig:3.2:Results-CREA-Abstracts-ClustersMetrics-DB}-\ref{fig:3.3:Results-CREA-Abstracts-ClustersMetrics-CH} show the evolution curves of the cluster validity metrics for medium and high strategies with $ \beta = 1.00 $ as a function of $ k $.
We observe that both the Silhouette score (Figure~\ref{fig:3.1:Results-CREA-Abstracts-ClustersMetrics-Silhouette}) and Calinski-Harabasz (Figure~\ref{fig:3.3:Results-CREA-Abstracts-ClustersMetrics-CH}) consistently favor the maximum value of $ k = 36 $, but this would result in an excessive number of topics.
The Davies-Bouldin index (Figure~\ref{fig:3.2:Results-CREA-Abstracts-ClustersMetrics-DB}) prefers more compact clusters, and favors the minimum $ k = 4 $.
We also observe a second good candidate suggested by the Davies-Bouldin index at around $ k = 16 $ approximately.
We will use this value for topic analysis, as it represents a good compromise for the number of topics given 40 documents, without being excessive.
We also note that the high strategy achieves better Silhouette and Davies-Bouldin scores, while the medium strategy performs better on the Calinski-Harabasz index.



The medium strategy for $ \beta = 1.00 $ and $ k = 16 $ on Abstracts (see in appendix Table~\ref{tab:X.4.A.1:FCA-Clusters-Med-B=1.00-K=16-Abstracts}) produces a dominant topic containing 270 terms, followed by 15 smaller clusters ranging from 11 to 21 terms, with relatively uniform sizes.
The generated labels help interpret these topics, particularly given the large number of terms involved. For the largest cluster, ChatGPT provides the label "\textit{Business Processes - Concepts, methods, and tools for modeling, managing, and automating business processes and enterprise systems}", which aligns with the researcher's areas of expertise.
However, the analysis remains difficult due to the overwhelming number of terms.

With the high strategy for $ \beta = 1.00 $ and $ k = 16 $ on Abstracts (see in appendix Table~\ref{tab:X.4.A.2:FCA-Clusters-Hig-B=1.00-K=16-Abstracts}), we can observe that the largest cluster contains only 24 terms, which represents significant filtering compared to the medium strategy.
Looking at the ChatGPT description, it provides the label "\textit{Systems Management - Covers organizational processes, governance, software, and tools for managing crises and business operations}".
We can see a loss of specificity compared to the previous "\textit{Business Processes}" label, as the description has become broader and more general.
This occurs because the terms that were previously grouped in the large cluster are now distributed across smaller clusters.
For example, terms such as \textit{EA framework}, \textit{business process}, and \textit{case management} that were previously grouped in the large cluster are now scattered across different smaller clusters.
With fewer terms available, ChatGPT attempts to generate coherent labels with the limited vocabulary at its disposal.

\subsubsection{Full Papers}
\label{subsubsection:4.2.2:Dataset-ResearchPapers-Results-FullPapers}

For the full papers, given the larger size of the corpus, we directly applied the high strategy in order to avoid generating an excessively large lattice.
In this setting, we tested $ \beta $ values ranging from $ 0.75 $ to $ 4.00 $, which corresponds to a range yet not explored.
The results of the high strategy on the Full Papers dataset, showing cluster sizes across different $ \beta $ and $ k $ values can be found in the appendix (see Table~\ref{tab:X.3.F.1:FCA-ClustersSize-Hig-K=8-FullPapers}, Table~\ref{tab:X.3.F.2:FCA-ClustersSize-Hig-K=16-FullPapers}, and Table~\ref{tab:X.3.F.3:FCA-ClustersSize-Hig-K=20-FullPapers}).

\begin{figure}[!ht]
    \centering
    \includegraphics[width=1.0\linewidth]{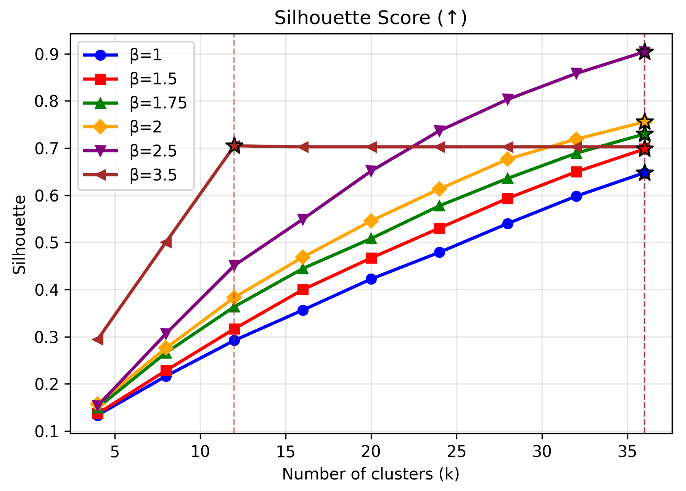}
    \caption{Silhouette score on clusters from Full Papers, with \textit{High Strategy} across $\beta $ and $ k $}
    \label{fig:4.1:Results-CREA-FullPapers-Silhouette}
\end{figure}
\begin{figure}[!ht]
    \centering
    \includegraphics[width=1.0\linewidth]{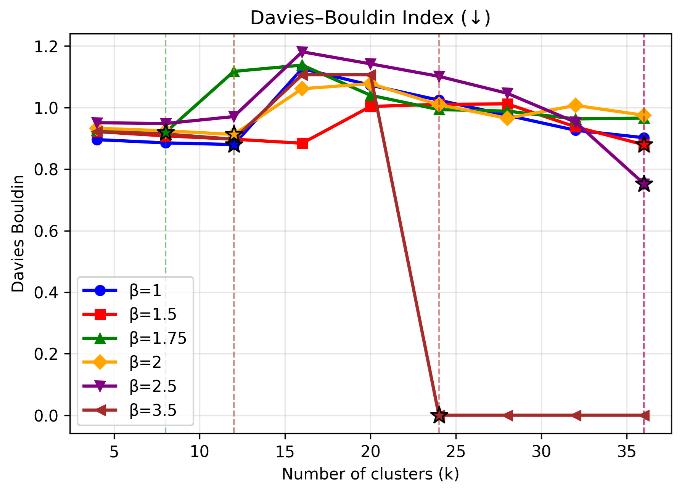}
    \caption{Davies-Bouldin index on clusters from Full Papers, with \textit{High Strategy} across $\beta $ and $ k $}
    \label{fig:4.2:Results-CREA-FullPapers-DB}
\end{figure}
\begin{figure}[!ht]
    \centering
    \includegraphics[width=1.0\linewidth]{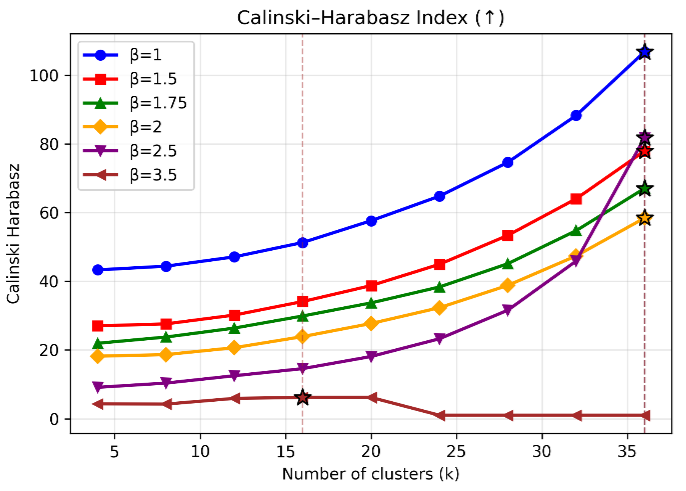}
    \caption{Calinski-Harabasz index on clusters from Full Papers, with \textit{High Strategy} across $\beta $ and $ k $}
    \label{fig:4.3:Results-CREA-FullPapers-CH}
\end{figure}

Figures~\ref{fig:4.1:Results-CREA-FullPapers-Silhouette}-\ref{fig:4.2:Results-CREA-FullPapers-DB}-\ref{fig:4.3:Results-CREA-FullPapers-CH} show the cluster metrics on Full Papers with high strategy, across $ \beta $ from $ 1 $ to $ 3.5 $, and varying the number $k$ of clusters from $ 4 $ to $ 36 $.
Among all strategies, Silhouette score (Figure~\ref{fig:4.1:Results-CREA-FullPapers-Silhouette}) consistently favors restrictive $ \beta $ values.
$ \beta = 2.5 $ systematically achieves good scores, reaching around $ 0.90 $ at $ k = 36 $.
$ \beta = 3.5 $ plateaus early around $ k = 12 $ because it becomes too restrictive, lacking sufficient material to form more clusters.
Davies-Bouldin index (Figure~\ref{fig:4.2:Results-CREA-FullPapers-DB}) shows the same problematic behavior for $ \beta = 3.5 $, which breaks down after $ k = 20 $.
This index tends to favor less restrictive $ \beta $ values, with better trends observed for $ \beta = 1.5 $.
For the Calinski-Harabasz index (Figure~\ref{fig:4.3:Results-CREA-FullPapers-CH}), the best scores are obtained for $ \beta = 1.0 $ and $ \beta = 1.5 $, reaching over $ 100 $ and $ \sim{}80 $ respectively at $ k = 36 $.
The more restrictive the $ \beta $ value, the worse the performance.
$ \beta = 2.5 $ stays below 50, and $ \beta = 3.5 $ collapses to nearly zero after $ k = 20 $.
These contradictory results across metrics highlight the challenge of parameter selection in CREA, with no single $ \beta $ value performing optimally across all evaluation criteria.

In the resulting clusters with $ k = 8 $ (CREA with high strategy and $ \beta = 1.00 $ in Table~\ref{tab:X.4.F.2:FCA-Clusters-Hig-B=2.50-K=8-FullPapers} and ChatGPT in Table~\ref{tab:X.5.F.2:LLM-Clusters-K=8-FullPapers}), we observe the same pattern as with abstracts: full papers reproduce the dominant cluster phenomenon.

Table~\ref{tab:XX:Results-FullPapers-CREA-S=H-B=2.50-k=8} presents the resulting clusters on Full Papers for CREA with high strategy ($ \beta = 1.00 $) using $ k = 8 $, and Table~\ref{tab:XX:Results-FullPapers-GPT-k=8} presents the equivalent 8 topics for GPT-5.
Other results for $ k = 8, 16, 20 $ are available in the~\hyperref[section:X:Appendix]{appendix section}.
In these results, we observe the same pattern as with Abstracts concerning CREA: a dominant cluster is always produced for $ \beta < 3.50 $.
Around $ \beta = 3.50 $, only 2 to 4 terms remain in each cluster, making the results meaningless.
While ChatGPT still attempts to provide labels, these descriptions become meaningless due to the severe vocabulary reduction.

A single dominant cluster still absorbs most of the terms, while increasing $ k $ only fragments it slightly without producing a more balanced distribution.
The largest cluster includes interesting terms like \textit{business process management}, \textit{collaboration}, \textit{enterprise modeling} and \textit{specification}, but remains too large to be interpreted.
Nevertheless, we can observe two patterns: (i) smaller clusters remain stable across different $k$, suggesting that CREA identifies structurally stable groupings of terms, and (ii) the dominant cluster fragments rather than distribute.
With the help of ChatGPT to interpret these larger clusters, we observe a progressive shift in labeling: at $k = 8$ the main cluster is labeled \textit{Methods \& Processes}, a broad category; at $k = 12$ it becomes \textit{Modeling Foundations}, more specific; and at $k = 20$ it narrows to \textit{Specification}, reflecting reduced vocabulary.
These observations characterize CREA's clustering behavior: increasing $k$ does not resolve the imbalance but instead subdivides the dominant cluster while preserving smaller, stable clusters.

Concerning GPT-5, the result is clear: we obtained well-balanced topics with ChatGPT, and the model strictly followed the instructions, consistently producing the requested number of topics with exactly five terms each.
Beyond structural compliance, the generated outputs show strong thematic coherence.
The topics align with the research domain, covering areas such as \textit{Enterprise Architecture} and \textit{Process Management}.
The labels are descriptive and informative, while the keywords appear specific and well chosen, making the topics straightforward to interpret.
When comparing Abstracts and Full Papers, we observe differences in stability and granularity.
Some topics, such as \textit{Enterprise Architecture} and \textit{Crisis Management}, appear consistently across both settings, while others shift or disappear altogether.
Abstracts produce more specialized topics, including \textit{AI Services and Business Rules}, whereas Full Papers yield broader and more academically oriented themes such as \textit{Workflow Automation} and \textit{Trust Models}.
This difference can be explained by the nature of abstracts, which are written to be concise and impactful, summarizing the paper in just a few lines and therefore emphasizing specific aspects.
In contrast, when provided with the entire document, GPT-5 captures the broader scope and recognizes that the contribution extends beyond a single specific angle.
At the same time, some overlap can be observed, with recurring terms such as \textit{trust}, \textit{modeling}, and \textit{alignment} appearing across multiple topics.
This reflects the central importance of these notions in the research domain, but it also shows that GPT-5 does not always enforce clear boundaries between themes. Depending on the intended use of the results, such overlap can either highlight robust thematic anchors or be seen as redundancy that reduces topic specificity.


\begin{center}
\begin{table*}[!ht]
    \centering
    \caption{Topics and their ChatGPT labels for CREA on Full Papers using \textit{High Strategy} ($\beta = 2.50$) and $k = 8$.}
    \begin{tabular}{ | c | L{14.0cm} | }
    \hline
    \cellcolor{black!15} Topic & \cellcolor{black!15} Label / Terms \\
    \hline
    \multirow{2}{*}{1}
    & \cellcolor{red!5} Methods \& Processes - Broad concepts, methods, and practices for designing, analyzing, and managing complex processes and systems \\ \cline{2-2}
    & \textit{ability, achieve, act, actors, add, al, algorithms, analyze, applicable, architecture, article, assess, authors, availability, available, behavior, best practice, business process management, change, collaboration, communication, community, company, complex, complexity, concept, concrete, context, contract, core, creation, criteria, decide, design process, desired, detailed, diagram, difficult, discuss, document, edges, effectiveness, emergence, enabling, enterprise modeling, entities, evaluation, evolution, executable, experience, experiment, expert, feature, fig, figurer, formal methods, fulfilled, goal, group, gs, hierarchy, higher, idea, illustrate, implementation, information, infrastructure, integration, interaction, life, literature, logic, main, manager, matrix, measurement, modeling languages, motivation, network, number, objectif, organizations, oriented, participant, present, problem, process, process design, process specification, project, proposer, react, regarding, related, rely, reputation, responsibilities, results, risk, science, selected, semantics, service, set, shared, skills, specification, state machine, states, success, successful, target, technique, technology, text, theory, time, treatment, type, valid, validation, value, version, way, workflow} \\ \hline
    \multirow{2}{*}{2}
    & \cellcolor{red!5} Healthcare Systems - Applications and documentation practices in engineering and healthcare contexts \\ \cline{2-2}
    & \textit{application, documentation, engineers, health, healthcare, phenomenon, reason, system} \\ \hline
    \multirow{2}{*}{3}
    & \cellcolor{red!5} Environmental Change - Handling changes, environments, and structured approaches to adaptation \\ \cline{2-2}
    & \textit{addition, changing, elaborate, environment, handling, head, high, structurer} \\ \hline
    \multirow{2}{*}{4}
    & \cellcolor{red!5} Conceptual Research - Conceptual and domain-oriented aspects of research and possible approaches \\ \cline{2-2}
    & \textit{conceptual, domain, existing, kind, possible, research, usually} \\ \hline
    \multirow{2}{*}{5}
    & \cellcolor{red!5} Business Strategy - Strategic business design, customer needs, and language in organizational contexts \\ \cline{2-2}
    & \textit{business, created, customer, designer, language, missing, strategic} \\ \hline
    \multirow{2}{*}{6}
    & \cellcolor{red!5} Process Improvement - Tools, strategies, and workshops for improving management systems and processes \\ \cline{2-2}
    & \textit{key, management systems, people, process improvement, strategy, tool, workshops} \\ \hline
    \multirow{2}{*}{7}
    & \cellcolor{red!5} Adaptive Management - Dynamic planning and definitions for management under variable conditions \\ \cline{2-2}
    & \textit{adapted, definition, dynamic, management, particularly, plan, weather} \\ \hline
    \multirow{2}{*}{8}
    & \cellcolor{red!5} Software \& Stakeholders - Software issues, stakeholder perspectives, and supply chain terminology \\ \cline{2-2}
    & \textit{lack, software, stakeholders, supply, supply chain, taxonomy, terms} \\ \hline
    \end{tabular}
    \label{tab:XX:Results-FullPapers-CREA-S=H-B=2.50-k=8}
\end{table*}
\end{center}

\begin{center}
\begin{table*}[!ht]
    \centering
    \caption{Topics and labels generated with ChatGPT from Full Papers ($k = 8$).}
    \begin{tabular}{ | c | L{14.0cm} | }
    \hline
    \cellcolor{black!15} Topic & \cellcolor{black!15} Label / Terms \\
    \hline
    \multirow{2}{*}{1}
    & \cellcolor{red!5} Enterprise Architecture - Frameworks and governance methods for aligning IT and organizational structures \\ \cline{2-2}
    & \textit{enterprise, architecture, governance, alignment, frameworks} \\ \hline
    \multirow{2}{*}{2}
    & \cellcolor{red!5} Process Management - Adaptive business process modeling and case management approaches \\ \cline{2-2}
    & \textit{business, process, case, management, adaptability} \\ \hline
    \multirow{2}{*}{3}
    & \cellcolor{red!5} Context Modeling - Use of ontologies, FCA, and reasoning for contextual representation \\ \cline{2-2}
    & \textit{context, modeling, ontology, FCA, reasoning} \\ \hline
    \multirow{2}{*}{4}
    & \cellcolor{red!5} Workflow Automation - Statecharts and BPMN for managing workflows and variability \\ \cline{2-2}
    & \textit{statecharts, workflows, BPMN, variability, automation} \\ \hline
    \multirow{2}{*}{5}
    & \cellcolor{red!5} Crisis Management - Simulation and resource planning for floods and emergency situations \\ \cline{2-2}
    & \textit{crisis, flood, emergency, simulation, resources} \\ \hline
    \multirow{2}{*}{6}
    & \cellcolor{red!5} Blockchain - Adoption and trust mechanisms in decentralized systems and contracts \\ \cline{2-2}
    & \textit{blockchain, adoption, trust, contracts, decentralization} \\ \hline
    \multirow{2}{*}{7}
    & \cellcolor{red!5} Trust Models - Social, digital, and technological dimensions of trust requirements \\ \cline{2-2}
    & \textit{trust, technology, digital, social, requirements} \\ \hline
    \multirow{2}{*}{8}
    & \cellcolor{red!5} Model-Driven Design - MDA approaches for aligning goals, features, and system models \\ \cline{2-2}
    & \textit{modeling, MDA, goals, features, alignment} \\ \hline
    \end{tabular}
    \label{tab:XX:Results-FullPapers-GPT-k=8}
\end{table*}
\end{center}



\section{\uppercase{Discussion}} 
\label{section:5:Discussion}

Three key aspects concerning the results between CREA and GPT-5 are now discussed.
First, \textit{transparency and reproducibility} is obviously well managed by CREA's deterministic and traceable steps, especially with the FCA at its core.
Each step is easily tunable and even replaceable, making it perfect for research purpose and explaining why specific topics emerges.
In contrast, the LLM operate as black box with an opaque reasoning process: a user can't trace \textit{why} and \textit{how} specific terms are grouped.
Proprietary models, by definition, do not share or list the data used for pre-training, or explain their tuning processes and architectural details.
Open source models only share the internal weighs or do not perform as well as proprietary ones.

Similarly, \textit{bias and ethics} concerns may exist in the proprietary models of LLM as nothing can be verified concerning the data used during the pre-training.
The output topics are inevitably influenced. 
In contrary, CREA acts closely to an unsupervised method relying only on the input documents.
The sole exception concerns BabelFy as it is a semantic network trained on data (but it can be replaced by any other method recognizing named entities).
We used publicly available data (research papers) in our experiments, but real-world applications may contain private data (like medical or corporate documents) that can be reused by the proprietary LLM following their terms of services.

Concerning the \textit{practical implementation}, CREA's modularity involves a certain complexity and multiple parameters across its pipeline: coherence thresholds for term filtering, binarization strategies with $ \beta $ coefficients, and clustering parameters including the optimal number of clusters.
Parameters cannot be decided a priori, and requires to rerun the steps for deciding which value produces the best results.
The numerous steps based on specific methods require the user to have a certain knowledge in the data science domain (like formal concept analysis and clustering).
In contrast, LLM are already widely used thanks to their ease of use.
They do not offer the same level of customization, but it allows to quickly prototype or explore new projects with minimal technical skill.
Multilingual support is available in both cases, however, in CREA's case, it is required to explicit the input language to BabelFy and even to configure the POS filtering for each language.

Subsequently, CREA is limited in its scalability: large datasets requires huge computational infrastructure because of the lattice exponential growth.
As long as the corpus remains in a correct size, the calculations can be achieved within hours.
LLM are obviously pre-trained with even bigger infrastructures, and suffer restrictions from their context window: large corpora requires to be separated within segments with the risk of losing global context and thematic coherence.
A similar limitation can be found in the current implementation of CREA with the use of BabelFy (except that BabelFy acts as a named entity linker and not as the main reasoning engine).


\section{\uppercase{Conclusion}} 
\label{section:6:Conclusion}

In this paper, we examine whether the CREA method could be applied beyond its initial educational context, and how it compares to prompt-based LLM for topic modeling.
The work was carried out on small, domain-specific datasets, not with the aim of proposing a definitive solution, but to explore the adaptability and limits of both methods.

Because of its modular design and its core based on FCA, CREA is ideal for traceability and reproducibility.
However, these advantages were overshadowed by practical challenges: heavy parameterization, high computational costs, and persistent imbalances in cluster sizes that rendered interpretation difficult.
These limitations hindered the use of classical coherence measures (which are themselves subject to debate in terms of validity).
The clustering-based evaluation metrics we adopted revealed CREA's limitations in generating balanced and easily interpretable topic structures.

On the other side, ChatGPT produced balanced and thematically coherent topics with remarkable ease.
However, its black-box design and reproducibility issues raise concerns about the interpretation of its results: transparency is valued in the academic community, thanks to the FAIR principles, as it allows to find possible bias or mistakes.

In conclusion, the choice of the topic modeling method should depend less on technical novelty and more on the intended use of the results, the need for transparency, and the constraints of the application context.
While our exploration revealed specific challenges in applying CREA to topic modeling, it contributes to our understanding of how different methodological approaches perform in practice and the importance of matching tools to their intended applications.

%% file: article_2_appendix_papers_list.tex


\begin{center}
\begin{table*}
    \small
    \noindent
    \caption{List of documents used in the Research Papers dataset (2019 - 2025).}
    \makebox[\textwidth]{
\begin{tabular}{ | >{\centering\let\newline\\\arraybackslash\hspace{0pt}}m{1.25cm} | L{17.00cm} | }
       \hline
       \rowcolor{black!15} ID & Reference (with DOI or URL) \\ \hline
       \hline
       2025 LR & H. Li and I. Rychkova, “Understanding Trust Formation in GPT Services: An Empirical Study,” RCIS 2025 Workshops and Research Projects Track. 20 - 23 May, 2025. Seville, Spain, 2025. [Online]. \newline
       \textit{URL: \href{https://ceur-ws.org/Vol-3987/paper2.pdf}{CEUR - Vol.3987 (Paper 2)}} \\ \hline
       \hline
       2025 AR & B. Aydogdu and I. Rychkova, “Exploring Trust in Blockchain Technology: A Critical Review of the Theoretical Acceptance Models,” in Proceedings of the 27th International Conference on Enterprise Information Systems, ICEIS 2025, Porto, Portugal, April 4-6, 2025, Volume 1, 2025, pp. 15–26. \newline
       \textit{DOI: \href{10.5220/0013143600003929}{10.5220/0013143600003929}} \\ \hline
       \hline
       2024 RKHPNS & I. Rychkova, E. K. Kambilo, N. Herbaut, O. Pastor, R. Noël, and C. Souveyet, “Technology-Aware Enterprise Modeling: Challenging the Model-Driven Architecture Paradigm,” in Enterprise, Business-Process and Information Systems Modeling - 25th International Conference, BPMDS 2024, and 29th International Conference, EMMSAD 2024, Limassol, Cyprus, June 3-4, 2024, Proceedings, 2024, vol. 511, pp. 388–396. \newline
       \textit{DOI: \href{10.1007/978-3-031-61007-3\_28}{10.1007/978-3-031-61007-3\_28}} \\
\hline
    \end{tabular}
    }
    \label{tab:X.REF.1:References-Papers-List-1}
\end{table*}
\end{center}


\begin{center}
\begin{table*}
    \small
    \noindent
    \caption{List of documents used in the Research Papers dataset (2014 - 2018).}
    \makebox[\textwidth]{
    \begin{tabular}{ | >{\centering\let\newline\\\arraybackslash\hspace{0pt}}m{1.25cm} | L{17.00cm} | }
       \hline
       \rowcolor{black!15} ID & Reference (with DOI or URL) \\ \hline
       \hline
       2018 RR & I. Rychkova and G. Regev, “From Goal-Achievement to the Maintenance of Relationships: Extending Business Process Models with Homeostasis and Appreciation,” in Proceedings of the 4th International Workshop on Socio-Technical Perspective in IS development (STPIS’18) co-located with 30th International Conference on Advanced Information Systems Engineering (CAiSE 2018), Tallinn, Estonia, June 12, 2018, 2018, vol. 2107, pp. 87–100. [Online]. \newline
       \textit{URL: \href{https://ceur-ws.org/Vol-2107/Paper7.pdf}{CEUR - Vol.2107 (Paper 7)}} \\ \hline
       \hline
       2017 ZR & J. Zdravkovic and I. Rychkova, “A Capability-Oriented Approach to IT Governance: The Case of Public Service Organizations,” in Advanced Information Systems Engineering Workshops - CAiSE 2017 International Workshops, Essen, Germany, June 12-16, 2017, Proceedings, 2017, vol. 286, pp. 39–49. \newline
       \textit{DOI: \href{10.1007/978-3-319-60048-2\_4}{10.1007/978-3-319-60048-2\_4}} \\ \hline
       \hline
       2017 RZ & I. Rychkova and J. Zdravkovic, “Towards Decentralized IT Governance in the Public Sector: A Capability-oriented Approach,” in Information Technology Governance in Public Organizations: Theory and Practice, L. Rusu and G. Viscusi, Eds. Cham: Springer International Publishing, 2017, pp. 107–132. \newline
       \textit{DOI: \href{10.1007/978-3-319-58978-7\_5}{10.1007/978-3-319-58978-7\_5}} \\
\hline
    \end{tabular}
    }
    \label{tab:X.REF.2:References-Papers-List-2}
\end{table*}
\end{center}


\begin{center}
\begin{table*}
    \small
    \noindent
    \caption{List of documents used in the Research Papers dataset (2010 - 2013).}
    \makebox[\textwidth]{
    \begin{tabular}{ | >{\centering\let\newline\\\arraybackslash\hspace{0pt}}m{1.25cm} | L{17.00cm} | }
       \hline
       \rowcolor{black!15} ID & Reference (with DOI or URL) \\ \hline
       \hline
       2013 RKPLG & I. Rychkova, M. Kirsch-Pinheiro, and B. L. Grand, “Context-Aware Agile Business Process Engine: Foundations and Architecture,” in Enterprise, Business-Process and Information Systems Modeling - 14th International Conference, BPMDS 2013, 18th International Conference, EMMSAD 2013, Held at CAiSE 2013, Valencia, Spain, June 17-18, 2013. Proceedings, 2013, vol. 147, pp. 32–47. \newline
       \textit{DOI: \href{10.1007/978-3-642-38484-4\_4}{10.1007/978-3-642-38484-4\_4}} \\ \hline
       \hline
       2013 TMRW & G. Tapandjieva, D. R. Marchetti, I. Rychkova, and A. Wegmann, “Towards the Definition, Implementation and Communication of an IT Strategy: The Case of IT Strategy at EPFL,” in Advanced Information Systems Engineering Workshops - CAiSE 2013 International Workshops, Valencia, Spain, June 17-21, 2013. Proceedings, 2013, vol. 148, pp. 99–110. \newline
       \textit{DOI: \href{10.1007/978-3-642-38490-5\_8}{10.1007/978-3-642-38490-5\_8}} \\ \hline
       \hline
       2013 BBRWa & B. Bajic-Bizumic, I. Rychkova, and A. Wegmann, “Simulation-Driven Approach for Business Rules Discovery,” in Advanced Information Systems Engineering Workshops - CAiSE 2013 International Workshops, Valencia, Spain, June 17-21, 2013. Proceedings, 2013, vol. 148, pp. 111–123. \newline
       \textit{DOI: \href{10.1007/978-3-642-38490-5\_9}{10.1007/978-3-642-38490-5\_9}} \\
\hline
    \end{tabular}
    }
    \label{tab:X.REF.3:References-Papers-List-3}
\end{table*}
\end{center}

%% file: article_3_appendix_metrics_clusters.tex


\begin{table*}[!ht]
    \centering
    \caption{Number of unique terms per strategy on Abstracts.}
    \begin{tabular}{ | c |c|c|c|c| }
        \hline
         \multirow{2}{*}{$ \beta $} &
        \multicolumn{4}{c|}{Strategy} \\
        \cline{2-5}
         & Direct & Medium & High & Low \\
        \hline
        $ 0.00 $ & \multirow{9}{*}{ $ 470 $ } & $ 295 $ & $ 175 $ & $ 175 $ \\
        \cline{1-1} \cline{3-5}
        $ 0.25 $ & & $ 347 $ & $ 175 $ & $ 175 $ \\ \cline{1-1} \cline{3-5}
        $ 0.50 $ & & $ 377 $ & $ 175 $ & $ 175 $ \\ \cline{1-1} \cline{3-5}
        $ 0.75 $ & & $ 386 $ & $ 175 $ & $ 169 $ \\ \cline{1-1} \cline{3-5}
        $ 1.00 $ & & $ 470 $ & $  91 $ & $  84 $ \\ \cline{1-1} \cline{3-5}
        $ 1.25 $ & & $ 470 $ & $  76 $ & $  29 $ \\ \cline{1-1} \cline{3-5}
        $ 1.50 $ & & $ 470 $ & $  51 $ & $   3 $ \\ \cline{1-1} \cline{3-5}
        $ 1.75 $ & & $ 470 $ & $  27 $ & $   0 $ \\ \cline{1-1} \cline{3-5}
        $ 2.00 $ & & $ 470 $ & $  17 $ & $   0 $ \\
        \hline
    \end{tabular}
    \label{tab:X.1.A.1:FCA-UniqueTerms-Abstracts}
\end{table*}

\begin{table*}[!ht]
    \centering
    \caption{Number of unique formal concepts per strategy on Abstracts.}
    \begin{tabular}{ | c |c|c|c|c| }
        \hline
         \multirow{2}{*}{$ \beta $} &
        \multicolumn{4}{c|}{Strategy} \\
        \cline{2-5}
         & Direct & Medium & High & Low \\
        \hline
        $ 0.00 $ & \multirow{9}{*}{ $ 558 $ } & $  43 $ & $ 119 $ & $ 191 $ \\
        \cline{1-1} \cline{3-5}
        $ 0.25 $ & & $  68 $ & $  87 $ & $ 144 $ \\ \cline{1-1} \cline{3-5}
        $ 0.50 $ & & $  98 $ & $  70 $ & $ 107 $ \\ \cline{1-1} \cline{3-5}
        $ 0.75 $ & & $ 153 $ & $  61 $ & $  63 $ \\ \cline{1-1} \cline{3-5}
        $ 1.00 $ & & $ 255 $ & $  49 $ & $  37 $ \\ \cline{1-1} \cline{3-5}
        $ 1.25 $ & & $ 361 $ & $  39 $ & $  20 $ \\ \cline{1-1} \cline{3-5}
        $ 1.50 $ & & $ 448 $ & $  29 $ & $   5 $ \\ \cline{1-1} \cline{3-5}
        $ 1.75 $ & & $ 500 $ & $  20 $ & $   5 $ \\ \cline{1-1} \cline{3-5}
        $ 2.00 $ & & $ 534 $ & $  14 $ & $   5 $ \\
        \hline
    \end{tabular}
    \label{tab:X.1.A.2:FCA-UniqueConcepts-Abstracts}
\end{table*}



\begin{table*}[!ht]
    \centering
    \caption{Number of unique terms and formal concepts per strategy on Full Papers.}
    \begin{tabular}{ | c |c|c|c|c| }
        \hline
        \multirow{4}{*}{$ \beta $} &
        \multicolumn{4}{c|}{Strategy} \\
        \cline{2-5}
         & Direct & \multicolumn{3}{c|}{High} \\
        \cline{2-5}
         & Terms & \multirow{2}{*}{Terms} & Excluded & \multirow{2}{*}{Concepts} \\
         & (Concepts) & & ($\%$) & \\
        \hline
        $ 0.75 $ & \multirow{14}{*}{ \shortstack{ $ 3110 $ \\ ($ 9535 $) } } & $ 1602 $ & $ 48.49 $ & $ 666 $ \\
        \cline{1-1} \cline{3-5}
        $ 1.00 $ & & $ 1040 $ & $ 66.56 $ & $ 406 $ \\ \cline{1-1} \cline{3-5}
        $ 1.25 $ & & $  906 $ & $ 70.87 $ & $ 263 $ \\ \cline{1-1} \cline{3-5}
        $ 1.50 $ & & $  625 $ & $ 79.90 $ & $ 175 $ \\ \cline{1-1} \cline{3-5}
        $ 1.75 $ & & $  453 $ & $ 85.43 $ & $ 126 $ \\ \cline{1-1} \cline{3-5}
        $ 2.00 $ & & $  343 $ & $ 88.97 $ & $  96 $ \\ \cline{1-1} \cline{3-5}
        $ 2.25 $ & & $  248 $ & $ 92.03 $ & $  71 $ \\ \cline{1-1} \cline{3-5}
        $ 2.50 $ & & $  176 $ & $ 94.34 $ & $  48 $ \\ \cline{1-1} \cline{3-5}
        $ 2.75 $ & & $  122 $ & $ 96.08 $ & $  44 $ \\ \cline{1-1} \cline{3-5}
        $ 3.00 $ & & $   78 $ & $ 97.49 $ & $  39 $ \\ \cline{1-1} \cline{3-5}
        $ 3.25 $ & & $   52 $ & $ 98.33 $ & $  32 $ \\ \cline{1-1} \cline{3-5}
        $ 3.50 $ & & $   37 $ & $ 98.81 $ & $  24 $ \\ \cline{1-1} \cline{3-5}
        $ 3.75 $ & & $   24 $ & $ 99.23 $ & $  19 $ \\ \cline{1-1} \cline{3-5}
        $ 4.00 $ & & $   15 $ & $ 99.52 $ & $  12 $ \\
        \hline
    \end{tabular}
    \label{tab:X.1.F.1:FCA-UniqueTermsConcepts-FullPapers}
\end{table*}




\begin{table*}[!ht]
    \centering
    \caption{Cluster metrics on Abstracts using \textit{Medium Strategy} ($\beta = 1.00$).}
    \label{tab:X.2.A.1:FCA-ClusterMetrics-Med-B=1.00-Abstracts}
\end{table*}


\begin{table*}[!ht]
    \centering
    \caption{Cluster metrics on Abstracts using \textit{High Strategy} ($\beta = 1.00$).}
    \label{tab:X.2.A.2:FCA-ClusterMetrics-Hig-B=1.00-Abstracts}
\end{table*}






\begin{table*}[!ht]
    \centering
    \caption{Cluster size on Abstracts across $ \beta $, with \textit{Medium Strategy} ($ k = 8 $).}
\begin{tabular}{ | c | c | c | c | }
        \hline
        \rowcolor{black!15} \multicolumn{4}{|c|}{High Strategy - $ k = 8 $} \\
        \hline
        \rowcolor{black!15} $ \beta $ & Largest & $2^{\text{nd}}$ cluster & Smallest \\
        \hline
		\cellcolor{black!15} $ 0.75 $ & $ 290 $ & $ 16 $ & $ 12 $ \\ \hline
		\cellcolor{black!15} $ 1.00 $ & $ 361 $ & $ 27 $ & $ 12 $ \\ \hline
		\cellcolor{black!15} $ 1.25 $ & $ 370 $ & $ 18 $ & $ 12 $ \\ \hline
    \end{tabular}
    \label{tab:X.3.A.1:FCA-ClustersSize-Med-K=8-Abstracts}
\end{table*}

\begin{table*}[!ht]
    \centering
    \caption{Cluster size on Abstracts across $ \beta $, with \textit{Medium Strategy} ($ k = 16 $).}
    \begin{tabular}{ | c | c | c | c | }
        \hline
        \rowcolor{black!15} \multicolumn{4}{|c|}{High Strategy - $ k = 16 $} \\
        \hline
        \rowcolor{black!15} $ \beta $ & Largest & $2^{\text{nd}}$ cluster & Smallest \\
        \hline
		\cellcolor{black!15} $ 0.75 $ & $ 202 $ & $ 16 $ & $ 10 $ \\ \hline
		\cellcolor{black!15} $ 1.00 $ & $ 270 $ & $ 21 $ & $ 11 $ \\ \hline
		\cellcolor{black!15} $ 1.25 $ & $ 247 $ & $ 34 $ & $ 10 $ \\ \hline
    \end{tabular}
    \label{tab:X.3.A.2:FCA-ClustersSize-Med-K=16-Abstracts}
\end{table*}

\begin{table*}[!ht]
    \centering
    \caption{Cluster size on Abstracts across $ \beta $, with \textit{Medium Strategy} ($ k = 20 $).}
    \begin{tabular}{ | c | c | c | c | }
        \hline
        \rowcolor{black!15} \multicolumn{4}{|c|}{High Strategy - $ k = 20 $} \\
        \hline
        \rowcolor{black!15} $ \beta $ & Largest & $2^{\text{nd}}$ cluster & Smallest \\
        \hline
		\cellcolor{black!15} $ 0.75 $ & $ 163 $ & $ 16 $ & $  9 $ \\ \hline
		\cellcolor{black!15} $ 1.00 $ & $ 196 $ & $ 44 $ & $ 10 $ \\ \hline
		\cellcolor{black!15} $ 1.25 $ & $ 210 $ & $ 34 $ & $  9 $ \\ \hline
    \end{tabular}
    \label{tab:X.3.A.3:FCA-ClustersSize-Med-K=20-Abstracts}
\end{table*}


\begin{table*}[!ht]
    \centering
    \caption{Cluster size on Abstracts across $ \beta $, with \textit{High Strategy} ($ k = 8 $).}
    \begin{tabular}{ | c | c | c | c | }
        \hline
        \rowcolor{black!15} \multicolumn{4}{|c|}{High Strategy - $ k = 8 $} \\
        \hline
        \rowcolor{black!15} $ \beta $ & Largest & $2^{\text{nd}}$ cluster & Smallest \\
        \hline
		\cellcolor{black!15} $ 0.75 $ & $ 122 $ & $ 11 $ & $  6 $ \\ \hline
		\cellcolor{black!15} $ 1.00 $ & $  55 $ & $  8 $ & $  4 $ \\ \hline
		\cellcolor{black!15} $ 1.25 $ & $  41 $ & $  7 $ & $  3 $ \\ \hline
    \end{tabular}
    \label{tab:X.3.A.4:FCA-ClustersSize-Hig-K=8-Abstracts}
\end{table*}

\begin{table*}[!ht]
    \centering
    \caption{Cluster size on Abstracts across $ \beta $, with \textit{High Strategy} ($ k = 16 $).}
    \begin{tabular}{ | c | c | c | c | }
        \hline
        \rowcolor{black!15} \multicolumn{4}{|c|}{High Strategy - $ k = 16 $} \\
        \hline
        \rowcolor{black!15} $ \beta $ & Largest & $2^{\text{nd}}$ cluster & Smallest \\
        \hline
		\cellcolor{black!15} $ 0.75 $ & $  73 $ & $ 11 $ & $  6 $ \\ \hline
		\cellcolor{black!15} $ 1.00 $ & $  24 $ & $  8 $ & $  4 $ \\ \hline
		\cellcolor{black!15} $ 1.25 $ & $  17 $ & $  7 $ & $  3 $ \\ \hline
    \end{tabular}
    \label{tab:X.3.A.5:FCA-ClustersSize-Hig-K=16-Abstracts}
\end{table*}

\begin{table*}[!ht]
    \centering
    \caption{Cluster size on Abstracts across $ \beta $, with \textit{High Strategy} ($ k = 20 $).}
    \begin{tabular}{ | c | c | c | c | }
        \hline
        \rowcolor{black!15} \multicolumn{4}{|c|}{High Strategy - $ k = 20 $} \\
        \hline
        \rowcolor{black!15} $ \beta $ & Largest & $2^{\text{nd}}$ cluster & Smallest \\
        \hline
		\cellcolor{black!15} $ 0.75 $ & $  57 $ & $ 11 $ & $  4 $ \\ \hline
		\cellcolor{black!15} $ 1.00 $ & $  16 $ & $  8 $ & $  2 $ \\ \hline
		\cellcolor{black!15} $ 1.25 $ & $   9 $ & $  7 $ & $  2 $ \\ \hline
    \end{tabular}
    \label{tab:X.3.A.6:FCA-ClustersSize-Hig-K=20-Abstracts}
\end{table*}



\begin{table*}[!ht]
    \centering
    \caption{Cluster size on Full Papers across $ \beta $, with \textit{High Strategy} ($ k = 8 $).}
    \begin{tabular}{ | c | c | c | c | }
        \hline
        \rowcolor{black!15} \multicolumn{4}{|c|}{High Strategy - $ k = 8 $} \\
        \hline
        \rowcolor{black!15} $ \beta $ & Largest & $2^{\text{nd}}$ cluster & Smallest \\
        \hline
		\cellcolor{black!15} $ 1.50 $ & $ 492 $ & $ 23 $ & $ 16 $ \\ \hline
		\cellcolor{black!15} $ 1.75 $ & $ 340 $ & $ 20 $ & $ 14 $ \\ \hline
		\cellcolor{black!15} $ 2.00 $ & $ 254 $ & $ 16 $ & $ 10 $ \\ \hline
		\cellcolor{black!15} $ 2.25 $ & $ 181 $ & $ 11 $ & $  8 $ \\ \hline
		\cellcolor{black!15} $ 2.50 $ & $ 125 $ & $  8 $ & $  7 $ \\ \hline
		\cellcolor{black!15} $ 2.75 $ & $  85 $ & $  6 $ & $  5 $ \\ \hline
		\cellcolor{black!15} $ 3.00 $ & $  49 $ & $  6 $ & $  3 $ \\ \hline
		\cellcolor{black!15} $ 3.25 $ & $  32 $ & $  4 $ & $  2 $ \\ \hline
		\cellcolor{black!15} $ 3.50 $ & $  19 $ & $  4 $ & $  2 $ \\ \hline
		\cellcolor{black!15} $ 3.75 $ & $   7 $ & $  4 $ & $  2 $ \\ \hline
		\cellcolor{black!15} $ 4.00 $ & $   6 $ & $  3 $ & $  2 $ \\ \hline
    \end{tabular}
    \label{tab:X.3.F.1:FCA-ClustersSize-Hig-K=8-FullPapers}
\end{table*}

\begin{table*}[!ht]
    \centering
    \caption{Cluster size on Full Papers across $ \beta $, with \textit{High Strategy} ($ k = 16 $).}
    \begin{tabular}{ | c | c | c | c | }
        \hline
        \rowcolor{black!15} \multicolumn{4}{|c|}{High Strategy - $ k = 16 $} \\
        \hline
        \rowcolor{black!15} $ \beta $ & Largest & $2^{\text{nd}}$ cluster & Smallest \\
        \hline
		\cellcolor{black!15} $ 1.50 $ & $ 370 $ & $ 23 $ & $ 14 $ \\ \hline
		\cellcolor{black!15} $ 1.75 $ & $ 216 $ & $ 39 $ & $ 11 $ \\ \hline
		\cellcolor{black!15} $ 2.00 $ & $ 167 $ & $ 16 $ & $  9 $ \\ \hline
		\cellcolor{black!15} $ 2.25 $ & $ 108 $ & $ 15 $ & $  7 $ \\ \hline
		\cellcolor{black!15} $ 2.50 $ & $  67 $ & $ 11 $ & $  6 $ \\ \hline
		\cellcolor{black!15} $ 2.75 $ & $  51 $ & $  6 $ & $  4 $ \\ \hline
		\cellcolor{black!15} $ 3.00 $ & $  29 $ & $  5 $ & $  2 $ \\ \hline
		\cellcolor{black!15} $ 3.25 $ & $  16 $ & $  4 $ & $  2 $ \\ \hline
		\cellcolor{black!15} $ 3.50 $ & $   4 $ & $  4 $ & $  2 $ \\ \hline
		\cellcolor{black!15} $ 3.75 $ & $   3 $ & $  2 $ & $  1 $ \\ \hline
		\cellcolor{black!15} $ 4.00 $ & $   3 $ & $  2 $ & $  1 $ \\ \hline
    \end{tabular}
    \label{tab:X.3.F.2:FCA-ClustersSize-Hig-K=16-FullPapers}
\end{table*}

\begin{table*}[!ht]
    \centering
    \caption{Cluster size on Full Papers across $ \beta $, with \textit{High Strategy} ($ k = 20 $).}
    \begin{tabular}{ | c | c | c | c | }
        \hline
        \rowcolor{black!15} \multicolumn{4}{|c|}{High Strategy - $ k = 20 $} \\
        \hline
        \rowcolor{black!15} $ \beta $ & Largest & $2^{\text{nd}}$ cluster & Smallest \\
        \hline
		\cellcolor{black!15} $ 1.50 $ & $ 301 $ & $ 30 $ & $ 13 $ \\ \hline
		\cellcolor{black!15} $ 1.75 $ & $ 188 $ & $ 20 $ & $  8 $ \\ \hline
		\cellcolor{black!15} $ 2.00 $ & $ 128 $ & $ 17 $ & $  7 $ \\ \hline
		\cellcolor{black!15} $ 2.25 $ & $  90 $ & $ 13 $ & $  6 $ \\ \hline
		\cellcolor{black!15} $ 2.50 $ & $  47 $ & $ 10 $ & $  5 $ \\ \hline
		\cellcolor{black!15} $ 2.75 $ & $  35 $ & $  6 $ & $  3 $ \\ \hline
		\cellcolor{black!15} $ 3.00 $ & $  21 $ & $  5 $ & $  2 $ \\ \hline
		\cellcolor{black!15} $ 3.25 $ & $  12 $ & $  4 $ & $  2 $ \\ \hline
		\cellcolor{black!15} $ 3.50 $ & $   4 $ & $  4 $ & $  2 $ \\ \hline
		\cellcolor{black!15} $ 3.75 $ & $   3 $ & $  2 $ & $  1 $ \\ \hline
		\cellcolor{black!15} $ 4.00 $ & $   3 $ & $  2 $ & $  1 $ \\ \hline
    \end{tabular}
    \label{tab:X.3.F.3:FCA-ClustersSize-Hig-K=20-FullPapers}
\end{table*}




\begin{center}
\begin{table*}[!ht]
    \noindent
    \caption{Topics and their ChatGPT labels for CREA on Abstracts using \textit{Medium Strategy} ($\beta = 1.00$) and $k = 16$.}
    \makebox[\linewidth]{
    }
    \label{tab:X.4.A.1:FCA-Clusters-Med-B=1.00-K=16-Abstracts}
\end{table*}
\end{center}

\begin{center}
\begin{table*}[!ht]
    \centering
    \caption{Topics and their ChatGPT labels for CREA on Abstracts using \textit{High Strategy} ($\beta = 1.00$) and $k = 16$.}
    \makebox[\linewidth]{
    \begin{tabular}{ | c | c | L{14.0cm} | }
    \hline
    \rowcolor{black!15} Topic & NbTerms & Label / Terms \\
    \hline
    \multirow{2}{*}{1} & \multirow{2}{*}{\textit{24}}
    & \cellcolor{red!5} Systems Management - Covers organizational processes, governance, software, and tools for managing crises and business operations \\ \cline{3-3}
    & & \textit{achieve, architecture, business, crisis management, criterion, decentralized organization, discuss, example, flood management, governance, importance, management, management process, offer, organization, process specification, semantics, service, software, specification, state, statecharts, system, tool} \\ \hline
    \multirow{2}{*}{2} & \multirow{2}{*}{\textit{8}}
    & \cellcolor{red!5} Enterprise Structure - Relates to concepts of education, enterprise architecture, and organizational structures \\ \cline{3-3}
    & & \textit{concept, education, enterprise architecture, organizational, organizational structure, propose, refer, size} \\ \hline
    \multirow{2}{*}{3} & \multirow{2}{*}{\textit{7}}
    & \cellcolor{red!5} Research Methods - Focuses on context, methodology, measurement, and information systems in research \\ \cline{3-3}
    & & \textit{context, experience, information, information system, measurement, methodology, research} \\ \hline
    \multirow{2}{*}{4} & \multirow{2}{*}{\textit{6}}
    & \cellcolor{red!5} Organizational Change - Addresses centralized styles, objectives, and changes within organizations \\
\hline
    \end{tabular}
    }
    \label{tab:X.4.A.2:FCA-Clusters-Hig-B=1.00-K=16-Abstracts}
\end{table*}
\end{center}



\begin{table*}[p!ht]
    \noindent
    \caption{Topics and their ChatGPT labels for CREA on Full Papers using \textit{High Strategy} ($\beta = 1.50$) and $k = 16$.}
    \makebox[\linewidth]{
    }
    \label{tab:X.4.F.1:FCA-Clusters-Hig-B=1.50-K=16-FullPapers}
\end{table*}

\begin{center}
\begin{table*}[!ht]
    \centering
    \caption{Topics and their ChatGPT labels for CREA on Full Papers using \textit{High Strategy} ($\beta = 2.50$) and $k = 8$.}
    \makebox[\linewidth]{
    \begin{tabular}{ | c | c | L{14.0cm} | }
    \hline
    \rowcolor{black!15} Topic & NbTerms & Label / Terms \\
    \hline
    \multirow{2}{*}{1} & \multirow{2}{*}{\textit{125}}
    & \cellcolor{red!5} Methods \& Processes - Broad concepts, methods, and practices for designing, analyzing, and managing complex processes and systems \\ \cline{3-3}
    & & \textit{ability, achieve, act, actors, add, al, algorithms, analyze, applicable, architecture, article, assess, authors, availability, available, behavior, best practice, business process management, change, collaboration, communication, community, company, complex, complexity, concept, concrete, context, contract, core, creation, criteria, decide, design process, desired, detailed, diagram, difficult, discuss, document, edges, effectiveness, emergence, enabling, enterprise modeling, entities, evaluation, evolution, executable, experience, experiment, expert, feature, fig, figurer, formal methods, fulfilled, goal, group, gs, hierarchy, higher, idea, illustrate, implementation, information, infrastructure, integration, interaction, life, literature, logic, main, manager, matrix, measurement, modeling languages, motivation, network, number, objectif, organizations, oriented, participant, present, problem, process, process design, process specification, project, proposer, react, regarding, related, rely, reputation, responsibilities, results, risk, science, selected, semantics, service, set, shared, skills, specification, state machine, states, success, successful, target, technique, technology, text, theory, time, treatment, type, valid, validation, value, version, way, workflow} \\ \hline
    \multirow{2}{*}{2} & \multirow{2}{*}{\textit{8}}
    & \cellcolor{red!5} Healthcare Systems - Applications and documentation practices in engineering and healthcare contexts \\ \cline{3-3}
    & & \textit{application, documentation, engineers, health, healthcare, phenomenon, reason, system} \\
\hline
    \end{tabular}
    }
    \label{tab:X.4.F.2:FCA-Clusters-Hig-B=2.50-K=8-FullPapers}
\end{table*}
\end{center}

\begin{center}
\begin{table*}[!ht]
    \centering
    \caption{Topics and their ChatGPT labels for CREA on Full Papers using \textit{High Strategy} ($\beta = 2.50$) and $k = 12$.}
    \makebox[\linewidth]{
    \begin{tabular}{ | c | c | L{14.0cm} | }
    \hline
    \rowcolor{black!15} Topic & NbTerms & Label / Terms \\
    \hline
    \multirow{2}{*}{1} & \multirow{2}{*}{\textit{98}}
    & \cellcolor{red!5} Modeling Foundations - Core principles, methods, and evaluation aspects of process and system modeling \\ \cline{3-3}
    & & \textit{ability, achieve, act, actors, al, algorithms, analyze, applicable, architecture, article, assess, authors, availability, available, behavior, best practice, change, communication, community, concept, concrete, context, contract, core, criteria, decide, desired, detailed, diagram, discuss, document, edges, effectiveness, emergence, enabling, entities, evaluation, evolution, executable, experiment, feature, figurer, fulfilled, goal, group, gs, higher, illustrate, implementation, information, infrastructure, integration, interaction, life, literature, logic, main, manager, matrix, measurement, motivation, network, number, organizations, oriented, participant, present, problem, process design, process specification, proposer, react, regarding, related, rely, reputation, responsibilities, results, risk, selected, semantics, set, skills, specification, state machine, states, target, technique, technology, text, theory, time, treatment, type, valid, value, version, way} \\ \hline
    \multirow{2}{*}{2} & \multirow{2}{*}{\textit{8}}
    & \cellcolor{red!5} Healthcare Systems - Applications of engineering and documentation in healthcare contexts \\ \cline{3-3}
    & & \textit{application, documentation, engineers, health, healthcare, phenomenon, reason, system} \\ \hline
    \multirow{2}{*}{3} & \multirow{2}{*}{\textit{8}}
    & \cellcolor{red!5} Environment Handling - Adaptation and structuring in changing environments \\ \cline{3-3}
    & & \textit{addition, changing, elaborate, environment, handling, head, high, structurer} \\
\hline
    \end{tabular}
    }
    \label{tab:X.4.F.3:FCA-Clusters-Hig-B=2.50-K=12-FullPapers}
\end{table*}
\end{center}

\begin{center}
\begin{table*}[!ht]
    \centering
    \caption{Topics and their ChatGPT labels for CREA on Full Papers using \textit{High Strategy} ($\beta = 2.50$) and $k = 16$.}
    \makebox[\linewidth]{
    \begin{tabular}{ | c | c | L{14.0cm} | }
    \hline
    \rowcolor{black!15} Topic & NbTerms & Label / Terms \\
    \hline
    \multirow{2}{*}{1} & \multirow{2}{*}{\textit{67}}
    & \cellcolor{red!5} Process Specification - Concepts, practices, and techniques for designing, specifying, and evaluating processes \\ \cline{3-3}
    & & \textit{achieve, act, actors, algorithms, applicable, article, availability, available, behavior, best practice, communication, community, concept, concrete, context, contract, core, decide, desired, diagram, discuss, document, edges, enabling, entities, evaluation, evolution, executable, experiment, feature, figurer, fulfilled, group, gs, infrastructure, interaction, life, literature, main, matrix, motivation, network, organizations, oriented, participant, present, process design, process specification, proposer, react, regarding, related, reputation, responsibilities, risk, selected, set, skills, specification, states, target, technique, text, time, treatment, type, value} \\ \hline
    \multirow{2}{*}{2} & \multirow{2}{*}{\textit{11}}
    & \cellcolor{red!5} Evaluation Criteria - Assessing effectiveness and validity using measurement, logic, and semantics \\ \cline{3-3}
    & & \textit{assess, criteria, effectiveness, information, logic, measurement, number, problem, semantics, state machine, valid} \\ \hline
    \multirow{2}{*}{3} & \multirow{2}{*}{\textit{9}}
    & \cellcolor{red!5} Scientific Analysis - Theoretical and technological analysis supported by research and results \\ \cline{3-3}
    & & \textit{ability, al, analyze, authors, higher, rely, results, technology, theory} \\ \hline
    \multirow{2}{*}{4} & \multirow{2}{*}{\textit{8}}
    & \cellcolor{red!5} Applied Systems - Applications in engineering, documentation, and healthcare systems \\
\hline
    \end{tabular}
    }
    \label{tab:X.4.F.4:FCA-Clusters-Hig-B=2.50-K=16-FullPapers}
\end{table*}
\end{center}

\begin{center}
\begin{table*}[!ht]
    \centering
    \caption{Topics and their ChatGPT labels for CREA on Full Papers using \textit{High Strategy} ($\beta = 2.50$) and $k = 20$.}
    \makebox[\linewidth]{
    \begin{tabular}{ | c | c | L{14.0cm} | }
    \hline
    \rowcolor{black!15} Topic & NbTerms & Label / Terms \\
    \hline
    \multirow{2}{*}{1} & \multirow{2}{*}{\textit{47}}
    & \cellcolor{red!5} Specification - Concepts related to defining, executing, and evaluating processes and systems \\ \cline{3-3}
    & & \textit{achieve, act, actors, algorithms, applicable, article, availability, behavior, best practice, communication, concrete, core, decide, desired, diagram, discuss, edges, enabling, entities, evolution, executable, feature, figurer, fulfilled, group, gs, infrastructure, interaction, life, literature, organizations, oriented, participant, process specification, react, regarding, related, reputation, risk, selected, skills, specification, states, technique, time, type, value} \\ \hline
    \multirow{2}{*}{2} & \multirow{2}{*}{\textit{10}}
    & \cellcolor{red!5} Process Design - Elements of designing processes within a given context \\ \cline{3-3}
    & & \textit{available, concept, context, present, process design, proposer, responsibilities, set, target, treatment} \\ \hline
    \multirow{2}{*}{3} & \multirow{2}{*}{\textit{9}}
    & \cellcolor{red!5} Theory - Analytical and theoretical aspects of technology and research \\ \cline{3-3}
    & & \textit{ability, al, analyze, authors, higher, rely, results, technology, theory} \\ \hline
    \multirow{2}{*}{4} & \multirow{2}{*}{\textit{8}}
    & \cellcolor{red!5} Healthcare Systems - Applications, documentation, and reasoning in healthcare engineering \\
\hline
    \end{tabular}
    }
    \label{tab:X.4.F.5:FCA-Clusters-Hig-B=2.50-K=20-FullPapers}
\end{table*}
\end{center}

\begin{center}
\begin{table*}[!ht]
    \centering
    \caption{Topics and their ChatGPT labels for CREA on Full Papers using \textit{High Strategy} ($\beta = 3.50$) and $k = 14$.}
    \makebox[\linewidth]{
    \begin{tabular}{ | c | c | L{14.0cm} | }
    \hline
    \rowcolor{black!15} Topic & NbTerms & Label / Terms \\
    \hline
    \multirow{2}{*}{1} & \multirow{2}{*}{\textit{4}}
    & \cellcolor{red!5} Strategy Tools - Key approaches and instruments used by people to support strategic planning \\ \cline{3-3}
    & & \textit{key, people, strategy, tool} \\ \hline
    \multirow{2}{*}{2} & \multirow{2}{*}{\textit{4}}
    & \cellcolor{red!5} Communication Change - The role of communication in adapting and creating new contexts \\ \cline{3-3}
    & & \textit{al, changing, communication, creation} \\ \hline
    \multirow{2}{*}{3} & \multirow{2}{*}{\textit{4}}
    & \cellcolor{red!5} Network Interaction - Enabling interactions and connections within a network \\ \cline{3-3}
    & & \textit{enabling, figurer, interaction, network} \\ \hline
    \multirow{2}{*}{4} & \multirow{2}{*}{\textit{3}}
    & \cellcolor{red!5} Complex Experience - Managing objectives in complex environments based on experience \\
\hline
    \end{tabular}
    }
    \label{tab:X.4.F.6:FCA-Clusters-Hig-B=3.50-K=14-FullPapers}
\end{table*}
\end{center}




\begin{center}
\begin{table*}[!ht]
    \centering
    \caption{Topics and labels with ChatGPT on Abstracts using $k = 5$.}
    \makebox[\linewidth]{
    \begin{tabular}{ | c | c | L{14.0cm} | }
    \hline
    \rowcolor{black!15} Topic & NbTerms & Label / Terms \\
    \hline
    \multirow{2}{*}{1} & \multirow{2}{*}{\textit{5}}
    & \cellcolor{red!5} Enterprise Architecture - Approaches to aligning organizational structures, governance, and decentralized systems \\ \cline{3-3}
    & & \textit{enterprise, architecture, governance, decentralization, alignment} \\ \hline
    \multirow{2}{*}{2} & \multirow{2}{*}{\textit{5}}
    & \cellcolor{red!5} Case Management - Modeling adaptable processes using declarative and state-oriented formalisms \\ \cline{3-3}
    & & \textit{case, management, statecharts, declarative, adaptability} \\ \hline
    \multirow{2}{*}{3} & \multirow{2}{*}{\textit{5}}
    & \cellcolor{red!5} Process Agility - Context-aware methods for dynamic, personalized, and adaptive process execution \\ \cline{3-3}
    & & \textit{process, agility, context-aware, mining, personalization} \\ \hline
    \multirow{2}{*}{4} & \multirow{2}{*}{\textit{5}}
    & \cellcolor{red!5} Blockchain Trust - Applying blockchain patterns to enhance trust and adoption in supply chains \\ \cline{3-3}
    & & \textit{blockchain, trust, patterns, supply, adoption} \\ \hline
    \multirow{2}{*}{5} & \multirow{2}{*}{\textit{5}}
    & \cellcolor{red!5} AI Services - Trust and awareness in the use of GPT-based AI systems within digital ecosystems \\ \cline{3-3}
    & & \textit{AI, GPT, services, awareness, ecosystems} \\ \hline
    \end{tabular}
    }
    \label{tab:X.5.A.1:LLM-Clusters-K=5-Abstracts}
\end{table*}
\end{center}

\begin{center}
\begin{table*}[!ht]
    \centering
    \caption{Topics and labels with ChatGPT on Abstracts using $k = 8$.}
    \makebox[\linewidth]{
    \begin{tabular}{ | c | c | L{14.0cm} | }
    \hline
    \rowcolor{black!15} Topic & NbTerms & Label / Terms \\
    \hline
    \multirow{2}{*}{1} & \multirow{2}{*}{\textit{5}}
    & \cellcolor{red!5} Enterprise Architecture - Methods for aligning organizational structures, governance, and decentralized systems \\ \cline{3-3}
    & & \textit{enterprise, architecture, governance, decentralization, alignment} \\ \hline
    \multirow{2}{*}{2} & \multirow{2}{*}{\textit{5}}
    & \cellcolor{red!5} Case Management - Declarative and state-based approaches for adaptable process modeling \\ \cline{3-3}
    & & \textit{case, management, statecharts, declarative, adaptability} \\ \hline
    \multirow{2}{*}{3} & \multirow{2}{*}{\textit{5}}
    & \cellcolor{red!5} Process Agility - Context-aware and personalized techniques for dynamic process execution \\ \cline{3-3}
    & & \textit{process, agility, context-aware, mining, personalization} \\ \hline
    \multirow{2}{*}{4} & \multirow{2}{*}{\textit{5}}
    & \cellcolor{red!5} Blockchain Trust - Blockchain patterns applied to trust and adoption in supply chains \\
\hline
    \end{tabular}
    }
    \label{tab:X.5.A.2:LLM-Clusters-K=8-Abstracts}
\end{table*}
\end{center}

\begin{center}
\begin{table*}[!ht]
    \centering
    \caption{Topics and labels with ChatGPT on Abstracts using $k = 12$.}
    \makebox[\linewidth]{
    \begin{tabular}{ | c | c | L{14.0cm} | }
    \hline
    \rowcolor{black!15} Topic & NbTerms & Label / Terms \\
    \hline
    \multirow{2}{*}{1} & \multirow{2}{*}{\textit{5}}
    & \cellcolor{red!5} Enterprise Architecture - Aligning governance and organizational structures in decentralized systems \\ \cline{3-3}
    & & \textit{enterprise, architecture, governance, decentralization, alignment} \\ \hline
    \multirow{2}{*}{2} & \multirow{2}{*}{\textit{5}}
    & \cellcolor{red!5} Case Management - Declarative and state-based approaches for adaptable process modeling \\ \cline{3-3}
    & & \textit{case, management, statecharts, declarative, adaptability} \\ \hline
    \multirow{2}{*}{3} & \multirow{2}{*}{\textit{5}}
    & \cellcolor{red!5} Process Agility - Context-aware execution frameworks for agile business processes \\ \cline{3-3}
    & & \textit{process, agility, context-aware, CAPE, execution} \\ \hline
    \multirow{2}{*}{4} & \multirow{2}{*}{\textit{5}}
    & \cellcolor{red!5} Blockchain Trust - Blockchain patterns applied to trust and adoption in supply chains \\
\hline
    \end{tabular}
    }
    \label{tab:X.5.A.3:LLM-Clusters-K=12-Abstracts}
\end{table*}
\end{center}



\begin{center}
\begin{table*}[!ht]
    \centering
    \caption{Topics and labels with ChatGPT on Full Papers using $k = 5$.}
    \makebox[\linewidth]{
    \begin{tabular}{ | c | c | L{14.0cm} | }
    \hline
    \rowcolor{black!15} Topic & NbTerms & Label / Terms \\
    \hline
    \multirow{2}{*}{1} & \multirow{2}{*}{\textit{5}}
    & \cellcolor{red!5} Enterprise Architecture - Frameworks and governance approaches for aligning IT and organizational structures \\ \cline{3-3}
    & & \textit{enterprise architecture, governance, alignment, decentralization, frameworks} \\ \hline
    \multirow{2}{*}{2} & \multirow{2}{*}{\textit{5}}
    & \cellcolor{red!5} Process Management - Adaptive modeling of business processes using case management and state-based methods \\ \cline{3-3}
    & & \textit{business process, case management, adaptability, declarative, statecharts} \\ \hline
    \multirow{2}{*}{3} & \multirow{2}{*}{\textit{5}}
    & \cellcolor{red!5} Context Modeling - Use of events, ontologies, and FCA for reasoning about contextual information \\ \cline{3-3}
    & & \textit{context modeling, events, ontology, FCA, reasoning} \\ \hline
    \multirow{2}{*}{4} & \multirow{2}{*}{\textit{5}}
    & \cellcolor{red!5} Blockchain - Governance and trust mechanisms in decentralized systems using smart contracts and DAOs \\ \cline{3-3}
    & & \textit{blockchain, DAO, governance, smart contracts, trust} \\ \hline
    \multirow{2}{*}{5} & \multirow{2}{*}{\textit{5}}
    & \cellcolor{red!5} Technology Acceptance - Predictors and mediating factors influencing adoption in acceptance models \\ \cline{3-3}
    & & \textit{technology acceptance, trust, predictors, mediators, SEM} \\ \hline
    \end{tabular}
    }
    \label{tab:X.5.F.1:LLM-Clusters-K=5-FullPapers}
\end{table*}
\end{center}

\begin{center}
\begin{table*}[!ht]
    \centering
    \caption{Topics and labels with ChatGPT on Full Papers using $k = 8$.}
    \makebox[\linewidth]{
    \begin{tabular}{ | c | c | L{14.0cm} | }
    \hline
    \rowcolor{black!15} Topic & NbTerms & Label / Terms \\
    \hline
    \multirow{2}{*}{1} & \multirow{2}{*}{\textit{5}}
    & \cellcolor{red!5} Enterprise Architecture - Frameworks and governance methods for aligning IT and organizational structures \\ \cline{3-3}
    & & \textit{enterprise, architecture, governance, alignment, frameworks} \\ \hline
    \multirow{2}{*}{2} & \multirow{2}{*}{\textit{5}}
    & \cellcolor{red!5} Process Management - Adaptive business process modeling and case management approaches \\ \cline{3-3}
    & & \textit{business, process, case, management, adaptability} \\ \hline
    \multirow{2}{*}{3} & \multirow{2}{*}{\textit{5}}
    & \cellcolor{red!5} Context Modeling - Use of ontologies, FCA, and reasoning for contextual representation \\ \cline{3-3}
    & & \textit{context, modeling, ontology, FCA, reasoning} \\ \hline
    \multirow{2}{*}{4} & \multirow{2}{*}{\textit{5}}
    & \cellcolor{red!5} Workflow Automation - Statecharts and BPMN for managing workflows and variability \\
\hline
    \end{tabular}
    }
    \label{tab:X.5.F.2:LLM-Clusters-K=8-FullPapers}
\end{table*}
\end{center}

\begin{center}
\begin{table*}[!ht]
    \centering
    \caption{Topics and labels with ChatGPT on Full Papers using $k = 12$.}
    \makebox[\linewidth]{
    \begin{tabular}{ | c | c | L{14.0cm} | }
    \hline
    \rowcolor{black!15} Topic & NbTerms & Label / Terms \\
    \hline
    \multirow{2}{*}{1} & \multirow{2}{*}{\textit{5}}
    & \cellcolor{red!5} Enterprise Architecture - Frameworks and governance methods for aligning IT and organizational structures \\ \cline{3-3}
    & & \textit{enterprise, architecture, governance, alignment, frameworks} \\ \hline
    \multirow{2}{*}{2} & \multirow{2}{*}{\textit{5}}
    & \cellcolor{red!5} Process Management - Adaptive modeling of business processes and case management \\ \cline{3-3}
    & & \textit{business, process, case, management, adaptability} \\ \hline
    \multirow{2}{*}{3} & \multirow{2}{*}{\textit{5}}
    & \cellcolor{red!5} Context Modeling - Ontologies and FCA for reasoning about contextual information \\ \cline{3-3}
    & & \textit{context, modeling, ontology, FCA, reasoning} \\ \hline
    \multirow{2}{*}{4} & \multirow{2}{*}{\textit{5}}
    & \cellcolor{red!5} Workflow Automation - Statecharts and BPMN for process variability and automation \\
\hline
    \end{tabular}
    }
    \label{tab:X.5.F.3:LLM-Clusters-K=12-FullPapers}
\end{table*}
\end{center}